%% file: main.tex
\documentclass[acmtog]{acmart}
\acmSubmissionID{1793}

\usepackage{booktabs} 

\usepackage{multirow}
\usepackage{colortbl}
\usepackage{cleveref}
\usepackage[table,dvipsnames]{xcolor}
\usepackage[most]{tcolorbox}
\usepackage{lipsum}
\usepackage{graphicx}
\usepackage{subcaption}
\usepackage{CJKutf8}

\crefname{figure}{Fig.}{Figs.}
\crefname{table}{Tab.}{Tabs.}
\crefname{section}{Sec.}{Secs.}
\crefname{appendix}{Appendix}{Appendices}

\usepackage[ruled]{algorithm2e} 

\SetAlFnt{\small}
\SetAlCapFnt{\small}
\SetAlCapNameFnt{\small}
\SetAlCapHSkip{0pt}

\copyrightyear{2025}
\acmYear{2025}
\setcopyright{acmlicensed}\acmConference[SA Conference Papers '25]{SIGGRAPH Asia 2025 Conference Papers}{December 15--18, 2025}{Hong Kong, Hong Kong}
\acmBooktitle{SIGGRAPH Asia 2025 Conference Papers (SA Conference Papers '25), December 15--18, 2025, Hong Kong, Hong Kong}
\acmDOI{10.1145/3757377.3763923}
\acmISBN{979-8-4007-2137-3/2025/12}
  
\begin{document}
\title{UTDesign: A Unified Framework for Stylized Text Editing and Generation in Graphic Design Images}

\author{Yiming Zhao}
\orcid{0009-0001-3011-2148}
\affiliation{%
 \institution{Wangxuan Institute of Computer Technology, Peking University}
 \city{Beijing}
 \country{China}}
\email{zhaoym@pku.edu.cn}

\author{Yuanpeng Gao}
\affiliation{%
 \institution{Wangxuan Institute of Computer Technology, Peking University}
 \city{Beijing}
 \country{China}}
\email{2200011612@stu.pku.edu.cn}

\author{Yuxuan Luo}
\affiliation{%
 \institution{Wangxuan Institute of Computer Technology, Peking University}
 \city{Beijing}
 \country{China}}
\email{2000017426@stu.pku.edu.cn}

\author{Jiwei Duan}
\affiliation{%
 \institution{Kingsoft Office}
 \city{Zhuhai}
 \country{China}}
\email{duanjiwei@wps.cn}

\author{Shisong Lin}
\affiliation{%
 \institution{Kingsoft Office}
 \city{Zhuhai}
 \country{China}}
\email{linshisong@wps.cn}

\author{Longfei Xiong}
\affiliation{%
 \institution{Kingsoft Office}
 \city{Zhuhai}
 \country{China}}
\email{xionglongfei@wps.cn}

\author{Zhouhui Lian}
\orcid{0000-0002-2683-7170}
\authornote{Corresponding author}
\affiliation{%
 \institution{Wangxuan Institute of Computer Technology, Peking University}
 \city{Beijing}
 \country{China}}
\affiliation{%
\institution{State Key Laboratory of General Artificial Intelligence, Peking University}
\city{Beijing}
\country{China}}
\email{lianzhouhui@pku.edu.cn}

\renewcommand\shortauthors{Y, Zhao. et al}

\begin{abstract}
AI-assisted graphic design has emerged as a powerful tool for automating the creation and editing of design elements such as posters, banners, and advertisements. While diffusion-based text-to-image models have demonstrated strong capabilities in visual content generation, their text rendering performance, particularly for small-scale typography and non-Latin scripts, remains limited. In this paper, we propose \textbf{UTDesign}, a unified framework for high-precision stylized text editing and conditional text generation in design images, supporting both English and Chinese scripts. Our framework introduces a novel DiT-based text style transfer model trained from scratch on a synthetic dataset, capable of generating transparent RGBA text foregrounds that preserve the style of reference glyphs. We further extend this model into a conditional text generation framework by training a multi-modal condition encoder on a curated dataset with detailed text annotations, enabling accurate, style-consistent text synthesis conditioned on background images, prompts, and layout specifications. Finally, we integrate our approach into a {fully automated} text-to-design (T2D) pipeline by incorporating pre-trained text-to-image (T2I) models and an MLLM-based layout planner. Extensive experiments demonstrate that \textbf{UTDesign} achieves state-of-the-art performance among open-source methods in terms of stylistic consistency and text accuracy, and also exhibits unique advantages compared to proprietary commercial approaches. {Code and data for this paper are available at \url{https://github.com/ZYM-PKU/UTDesign}}.
\end{abstract}

%
%

%


\begin{teaserfigure}
  \centering
  \includegraphics[width=\textwidth]{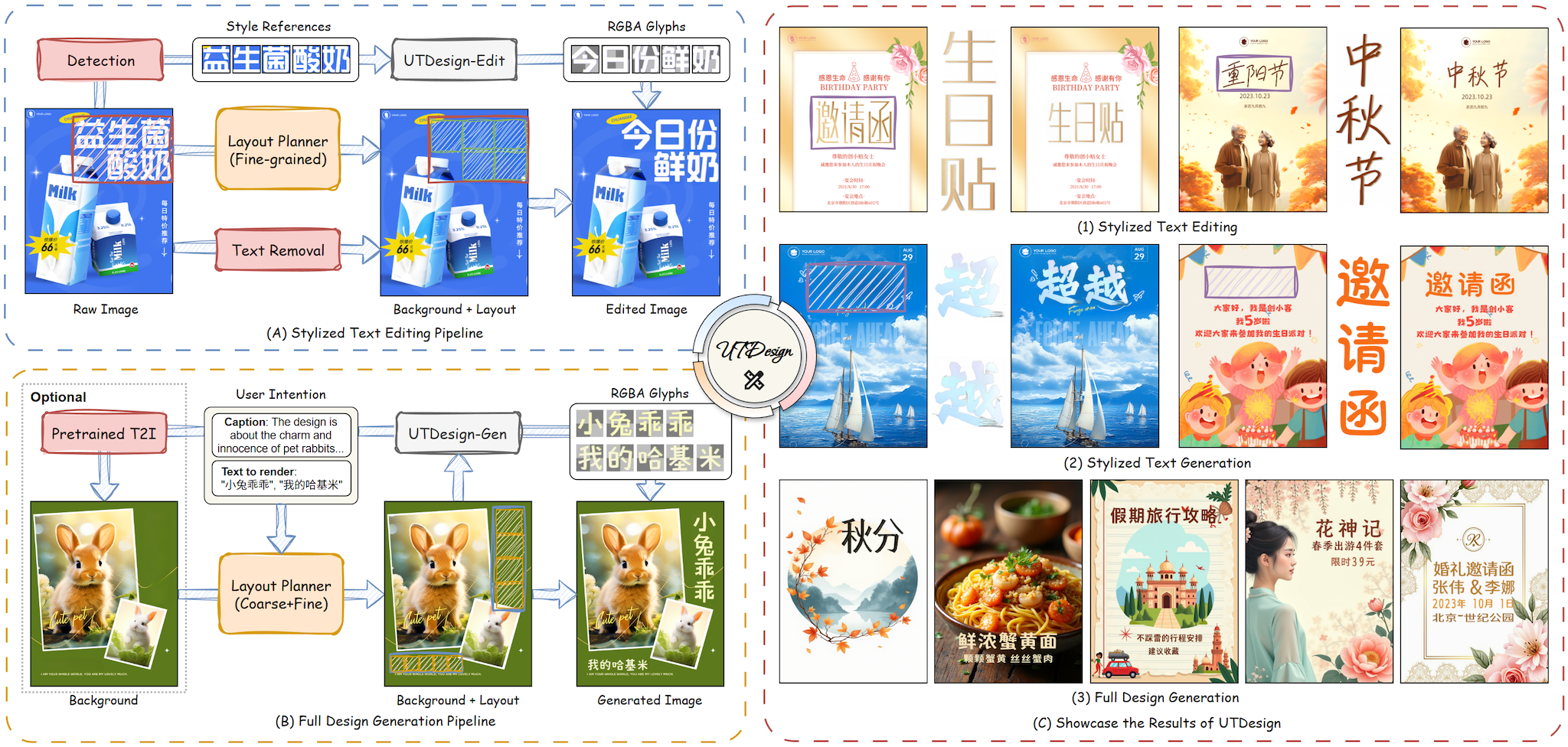}
  \caption{\textbf{UTDesign} supports editing arbitrary stylized text in design images (A) as well as generating complete design images (B). On the left side, we illustrate the pipeline for the two tasks, while the right side showcases the results of \textbf{UTDesign} across three different applications: (1) stylized text editing, (2) conditional stylized text generation, and (3) full design image generation.}
  \label{fig:teaser}
\end{teaserfigure}

\maketitle

\input{sec/1_intro}
\input{sec/2_related}

\input{sec/3_method}
\input{sec/4_experiment}
\input{sec/5_conclusion}

\bibliographystyle{ACM-Reference-Format}
\bibliography{main}

\input{appendix}

\end{document}

%% file: sec/1_intro.tex
\section{Introduction}
AI-assisted graphic design is an emerging application domain that leverages advanced AI techniques, particularly image generation models, to automate the creation and editing of visual elements such as posters, banners, and advertisements. A typical automated design system consists of a comprehensive pipeline encompassing multiple sub-tasks, including background generation, layout planning, and stylized text synthesis. Among these, artistic text rendering remains particularly challenging due to its critical role in determining the visual quality and aesthetic appeal of the final output.

Recent advances in diffusion-based image generation, especially text-to-image (T2I) models~\cite{chen2024pixart, esser2024scaling, team2024kolors, zheng2024cogview3, liu2024playground} built on Diffusion Transformers (DiTs)~\cite{peebles2023scalable}, have demonstrated impressive capabilities in generating high-quality, diverse design images. However, these models still struggle with accurate text rendering, particularly for small-scale typography and non-Latin scripts such as Chinese. Moreover, existing approaches often lack fine-grained editing capabilities while preserving stylistic consistency, making them unsuitable for real-world scenarios requiring precise modification of existing text elements.

To address these challenges, we propose a unified framework for high-precision stylized text editing and generation in design images, supporting both English and Chinese scripts, as illustrated in~\cref{fig:teaser}. Our method begins by training a DiT-based text style transfer model from scratch on a synthetic dataset, enabling accurate editing while preserving stylistic consistency. Given an arbitrary number of style-reference glyphs and the target text content, the model generates stylized text that consistently preserves the font characteristics and texture of the references. Together with a transparency glyph VAE decoder, our method enables the output of text foregrounds in RGBA format, facilitating seamless integration with existing scene-text detection techniques for {stylized} text editing in design contexts. 

Building upon this, we further extend the DiT model into a conditional text generation framework tailored for design images. To support this, we construct a dedicated dataset containing full design images, textual descriptions, and fine-grained text annotations specifying text content and spatial coordinates. We train a multi-modal condition encoder that takes as input the background image, prompt with the located target text and outputs features aligned with reference glyph styles. Integrating this encoder into the DiT backbone, along with post-training refinement, enables coherent and accurate stylized text generation under diverse visual conditions.

Finally, we develop a {fully automated} text-to-design (T2D) pipeline by integrating our DiT model with pre-trained T2I models and a layout planner guided by multi-modal large language models (MLLMs)~\cite{liu2023visual, li2024llava, bai2023qwen, bai2025qwen2, wang2024qwen2, chen2024internvl, chen2024expanding}. The pipeline translates user-provided textual descriptions into complete graphic designs through a sequence of stages: background generation, layout planning, and stylized text rendering. System-level evaluations show that our framework produces visually coherent and aesthetically compelling results, outperforming existing open-source methods and approaching the quality of commercial tools. In particular, the native support for transparent RGBA text foregrounds significantly enhances flexibility in practical editing workflows.

Our main contributions include:
\begin{enumerate}
    \item We propose \textbf{UTDesign}, a novel unified DiT-based framework for stylized text editing and generation in design images. Equipped with a customized transparency glyph VAE decoder, our system produces high-quality transparent text foregrounds. Extensive experiments show that \textbf{UTDesign} achieves state-of-the-art performance in both text rendering accuracy and stylistic consistency.
    \item We construct a large-scale synthetic dataset for training text editing models, and curate a real-world design image dataset with fine-grained text annotations, which is extendable for training layout planning and text synthesis models.
    \item We integrate \textbf{UTDesign} with T2I models and an MLLM-based layout planner, forming a {fully automated} T2D system that translates user intentions into finalized graphic designs, demonstrating strong potential for real-world applications. 
\end{enumerate}

%% file: sec/2_related.tex
\section{Related Work}

\subsection{Visual Text Rendering in Image Generation}
The relatively limited text rendering capability of UNet-based diffusion models has long been recognized as a major limitation.
A variety of methods have been proposed to address this issue by improving model architectures and training objectives.
Some approaches~\cite{yang2023glyphcontrol, zhang2024brush} incorporate ControlNet~\cite{zhang2023adding} and glyph reference images to provide stronger conditional guidance, enabling the base model to generate more accurate visual text.
Other methods~\cite{chen2024textdiffuser, liu2024glyph} enhance text structure understanding of text encoders by employing character-level tokenization.
Meanwhile, several works~\cite{zhao2024udifftext, wang2025designdiffusion} introduce well-crafted loss functions to further improve text rendering quality.
Subsequent research~\cite{tuo2024anytext, tuo2024anytext2, ma2025glyphdraw2, liu2024glyphv2} combines these advancements to develop models capable of accurate text rendering in both English and Chinese.
While these methods significantly improve the text rendering performance of UNet-based diffusion models, they often compromise the overall aesthetic quality, which makes them less suitable for real-world graphic design applications.

With the advent of DiT-based models (e.g., Stable Diffusion 3~\cite{esser2024scaling}, Flux~\footnote{FLUX.1-dev: \url{https://huggingface.co/black-forest-labs/FLUX.1-dev}}), the text rendering capability of T2I methods has markedly improved.
Recent efforts have focused on directly enhancing these models' performance in text-heavy or complex scenarios through attention manipulation~\cite{du2025textcrafter}, data augmentation~\cite{zhao2025lex}, and advanced sampling algorithms~\cite{hu2024amo}.
In particular, Seedream 2.0~\cite{gong2025seedream} incorporates several optimizations for text rendering, including enhanced datasets, multi-encoder architectures, and post-training strategies, achieving high-quality text rendering. More recently, models such as GPT-4o-image and Seedream 3.0~\cite{gao2025seedream3} have demonstrated strong capabilities in multi-lingual text rendering.
However, there remains a lack of open-source models that achieve comparable performance to these state-of-the-art proprietary systems.

\subsection{System-level Approaches for Auto Graphic Design}
Several works have attempted to build a complete pipeline for system-level graphic design automation.
AutoPoster~\cite{lin2023autoposter} generates posters through a four-stage process that leverages both product images and titles.
PosterMaker~\cite{gao2025postermaker} integrates background generation and visual text rendering into a unified framework, achieving accurate Chinese text rendering. However, its aesthetic performance remains limited.
COLE~\cite{jia2023cole} and OpenCOLE~\cite{inoue2024opencole} leverage open-source T2I models to generate background images, followed by layout planning and text style design using MLLMs. The planned text is then rendered onto the background to produce the final output.
POSTA~\cite{chen2025posta} employs an inpainting-based method to apply stylized textures to the rendered text, resulting in more visually appealing outcomes. Nevertheless, these approaches primarily focus on layout planning rather than text generation, limiting their ability to support designs with previously unseen fonts.
ART~\cite{pu2025art} introduces a novel and effective multi-layered approach for generating high-quality, editable design images, but it does not support non-Latin languages due to limitations in its base model.

%% file: sec/3_method.tex
\section{Method}

\begin{figure*}[!t]
\begin{minipage}[!t]{1\linewidth}
\centering
\includegraphics[width=1\textwidth]{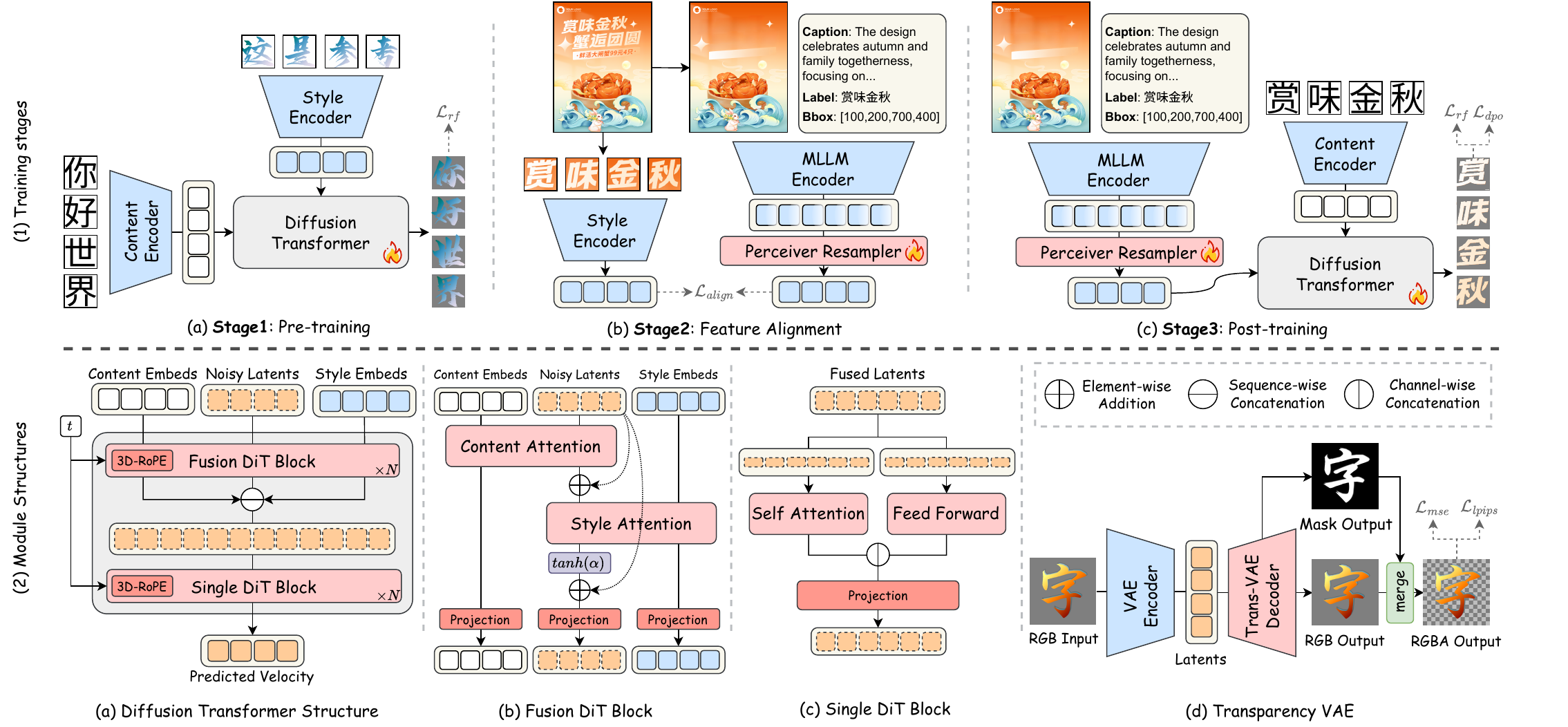}
\end{minipage}
\caption{Overview of the proposed \textbf{UTDesign}. The first row illustrates the training stages of our model, including: Stage1 (1a): Train from scratch a DiT with content/style encoders to conduct style-preserved text editing; Stage2 (1b): Extract guidance condition from the design background and textual description using MLLM encoder and align the encoded features with the pre-trained style encoder; Stage3 (1c): Replace the style encoder with the MLLM encoder and form a conditional glyph generation model through post-training. The second raw illustrates the detailed structure of the proposed DiT (2a,2b,2c), and shows the training process of our transparency glyph VAE (2d).}
\label{fig:method}
\end{figure*}

\subsection{Overview}
\label{sec: overview}

{Since training a single, unified model to} perform both precise editing and conditional generation on diverse real-world data presents a significant challenge, we introduce a progressive training strategy to ensure that the model first learns a disentangled representation of glyph content and style before it can master the application of those styles within complex conditional contexts.

As illustrated in~\cref{fig:method}, our approach is built upon a DiT architecture and consists of three training stages. In the first stage, we train the DiT model from scratch on synthetic data, incorporating a {glyph content encoder} and style encoder to enable style-preserving text editing.
In the second stage, we introduce an MLLM-based condition encoder, trained on a proposed design text image dataset, to extract guidance conditions from both the design background and textual descriptions. The encoded features are aligned with those from the glyph style encoder to ensure a consistent representation space.
In the third stage, we replace the style encoder with the condition encoder and conduct post-training, resulting in a conditional text generation model. By integrating a transparency glyph VAE and an MLLM-based layout planner, we construct a complete pipeline for stylized text editing and generation in design images, as shown in~\cref{fig:teaser}.
The following sections detail each training stage and the key components of our system.

\subsection{Data Curation}

\begin{figure}[!t]
\begin{minipage}[!t]{1\linewidth}
\centering
\includegraphics[width=1\textwidth]{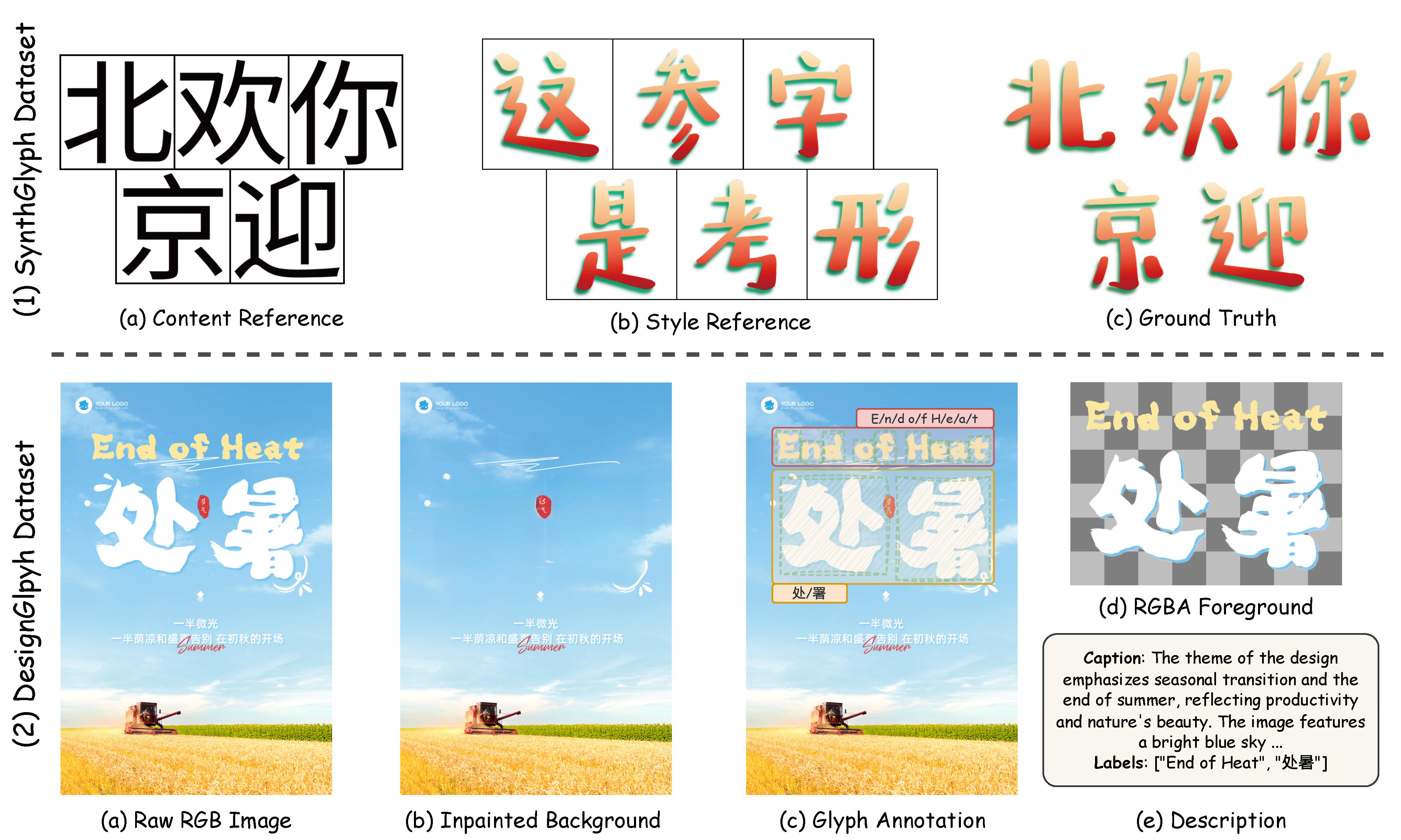}
\end{minipage}
\caption{Illustration of some examples of our proposed datasets.}
\label{fig:dataset}
\end{figure}


To train a unified model for text editing and generation, we constructed a large-scale synthetic stylized glyph dataset (\textbf{SynthGlyph Dataset}) and a design text image dataset (\textbf{DesignText Dataset}) with corresponding design images and fine-grained annotations, as detailed below.

\subsubsection{SynthGlyph Dataset}

To facilitate style transfer across arbitrary font styles, we collected a set of 4,194 fonts in TrueType format (TTF) and rendered 6,857 different characters for each font. The characters include Chinese glyphs defined by the GB6763 standard, as well as 94 English letters and symbols. This process resulted in approximately 28.8M stylized character instances. Additionally, we designed a glyph augmentation pipeline to render stylized text with varying colors and textures, further enhancing the diversity of the dataset. As shown in~\cref{fig:dataset} (1), when sampling from the dataset, we obtain a triplet $(\mathcal{R}_{c}, \mathcal{R}_{s}, \mathcal{GT})$, where $\mathcal{R}_{c}$ and $\mathcal{R}_{s}$ represent the content and style reference images, respectively, and $\mathcal{GT}$ denotes the ground truth in RGBA format rendered by applying the style of $\mathcal{R}_{s}$ to the content of $\mathcal{R}_{c}$. During style transfer training, we apply perturbations to $\mathcal{R}_{s}$, including Gaussian blur, down-sampling, random noise addition, and random background changes, to further improve the model's generalization ability.

\subsubsection{DesignText Dataset}


To further train a text synthesis model applicable to real-world graphic design scenarios, we curated a stylized text dataset based on real-world design images. This dataset consists of 115.5k diverse graphic design samples, from variate data sources. To handle datasets with varying levels of annotation granularity, we designed a fully automated annotation processing pipeline, incorporating text detection and recognition, foreground extraction, and image captioning. Please refer to Sec. 2 of the supplementary material for more details. 
As illustrated in~\cref{fig:dataset} (2), each data sample includes a design image, an extracted background image, a text description, the text content with glyph-level bounding box annotations and the foreground text in RGBA format, providing rich references for training different models.

\subsection{Stage 1: Pre-training of the Editing Model}


As shown in~\cref{fig:method} (1a), our editing model comprises a DiT-based backbone, a glyph content encoder $\mathcal{C}$, and a glyph style encoder $\mathcal{S}$. The content encoder is a ViT based on the pre-trained DINOv2~\cite{oquab2024dinov2} that can effectively capture the structural characteristics of glyphs, while the style encoder adopts the weights of pre-trained CLIP~\cite{radford2021learning} to extract stylistic features. Each encoder is followed by a projector composed of normalized transformer blocks, which maps the extracted features into a shared normalized latent space to facilitate downstream learning within the DiT backbone. During training, we jointly optimize the parameters of the DiT backbone and the two projectors.

In each training step, we sample a fixed-resolution batch from the \textbf{SynthGlyph Dataset}, including content reference glyphs $\mathcal{R}c = \{r_{c1}, r_{c2}\dots r_{cn}\}$, style reference glyphs $\mathcal{R}s = \{r_{s1}, r_{s2}\dots r_{sm}\}$, and corresponding ground truth glyphs $\mathcal{GT} = \{gt_1, gt_2\dots gt_n\}$, which are generated by applying the sampled styles to the content references. Following the Rectified Flow (RF)~\cite{esser2024scaling} framework, we denote $x_1$ as the VAE-encoded latent of $\mathcal{GT}$, $x_0 \sim \mathcal{N}(0, I)$ as a Gaussian noise sample, and $t \in [0, 1]$ as a randomly sampled timestep. The parameterized model $v_\theta$ takes the noisy latent $x_t$ as input and predicts the velocity field conditioned on both $\mathcal{R}_c$ and $\mathcal{R}_s$. The training objective is defined as:

\begin{small}
\begin{gather}
    x_t = tx_1+(1-t)x_0,\\
    v_t = \frac{dx_t}{dt}=x_1-x_0,\\
    \mathcal{L}_{rf}=\mathbb{E}_{x_0,x_1,t,\mathcal{R}_c,\mathcal{R}_s}||v_{\theta}(x_t,t,\mathcal{R}_c,\mathcal{R}_s)-v_t||^2_2.
\end{gather}
\end{small}

As depicted in~\cref{fig:method} (2a), we concatenate the content and style embeddings extracted by the encoders with the noisy latents and feed the sequence into the DiT. The network includes a combination of fusion DiT blocks and single DiT blocks to predict the target velocity field. We adopt 3D-RoPE to distinguish both the token type and its corresponding character identity and more details are provided in Sec. 3 of the supplementary material.

As illustrated in~\cref{fig:method} (2b,2c), within the fusion DiT blocks, the noisy latents attend to both content and style embeddings via full attention and are updated accordingly. A tanh gating mechanism is introduced to control the strength of style injection. The single DiT blocks take the full concatenated latent sequence as input and apply parallel self-attention and feedforward modules, followed by MLP projection, enabling global modeling while maintaining computational efficiency. To ensure stable training, we use RMSNorm across all attention operations.

After the pre-training, we fine-tune the whole model on the \textbf{DesignText Dataset} to extend the editing capabilities of the model to real design cases. The specific model design and training strategy enable our system to flexibly accept arbitrary numbers of content and style references, achieving consistent text editing with effective disentanglement between content and style.

\subsection{Stage 2: Feature Alignment of the Encoders}
To fully exploit the disentangled glyph style feature space learned in Stage 1 and extend the text editing model to a conditional text generation model, we introduce a multi-modal condition encoder based on an MLLM, along with a feature alignment mechanism. Specifically, during training, we sample a triplet $(\mathcal{R}_s, \mathcal{B}, \mathcal{D})$ from the \textbf{DesignText Dataset}, where $\mathcal{R}_s$ denotes the style reference, $\mathcal{B}$ is the background image of the design, and $\mathcal{D}$ contains textual description information, including an image caption, the target text to render, and its bounding box position.

We feed the design background and textual description into a frozen MLLM $\mathcal{M}$ using a carefully crafted instruction prompt, and extract the final hidden states as multi-modal condition features. Given the variable token length of MLLM outputs across different inputs, we employ a Perceiver Resampler~\cite{alayrac2022flamingo} module $\mathcal{P}$ to convert these variable-length embeddings into a fixed-size representation. The Perceiver Resampler uses a set of learnable query tokens to attend over the MLLM outputs and perform feature projection, enabling consistent downstream processing. Further implementation details are provided in the supplementary material.

As shown in~\cref{fig:method} (1b), we freeze both the style encoder and the MLLM during training, and optimize only the parameters of the Perceiver Resampler. For simplicity and stability, we employ an L2 loss to align the projected condition features with the learned style embedding space, formally defined as:
\begin{small}
\begin{equation}            
\mathcal{L}_{align}=\mathbb{E}_{\mathcal{R}_s, \mathcal{B}, \mathcal{D}}||\mathcal{P}_{\theta}(\mathcal{M}(\mathcal{B}, \mathcal{D}))-\mathcal{S}(\mathcal{R}_s)||^2_2.
\end{equation}
\end{small}

\subsection{Stage 3: Post-training of the Generation Model}
As shown in~\cref{fig:method} (1c), we replace the style encoder from Stage 1 with the MLLM encoder trained in Stage 2, and perform post-training using both Supervised Fine-Tuning (SFT) and Direct Preference Optimization (DPO), resulting in a conditional text generation model guided by background images and textual descriptions. Specifically, we first filter a high-quality and diverse subset from the \textbf{DesignText Dataset} and fine-tune the DiT using LoRA with a Rectified Flow loss, similar to Stage 1. See Sec. 2 of the supplementary materials for more details about the data filtering strategy.

To further enhance the quality and diversity of the generated images, we employ the Reinforcement Learning from Human Feedback (RLHF) mechanism for diffusion models. Specifically, for each training instance, we use the fine-tuned model to generate $k$ candidate samples, which are then scored and ranked by a pre-trained aesthetic reward model~\cite{kirstain2023pick}. Based on these rankings, we construct win–lose pairs for further training. Following the Diffusion-DPO framework~\cite{wallace2024diffusion}, we formulate the Rectified Flow-based objective to encourage the model to favor higher-ranked outputs, which is defined as:
\begin{small}
\begin{gather}
    x^w_t = tx^w_1+(1-t)x^w_0,\quad x^l_t = tx^l_1+(1-t)x^l_0,\\
    v^w_t = \frac{dx^w_t}{dt}=x^w_1-x^w_0,\quad v^l_t = \frac{dx^l_t}{dt}=x^l_1-x^l_0,\\
    \begin{split}
        &\mathcal{L}_{dpo} =  - \mathbb{E}_{x^w_0, x^l_0, x^w_1, x^w_1, t, \mathcal{B}, \mathcal{D}, \mathcal{R}_c, \mathcal{R}_s}\left[\log \sigma \left( - \mathcal{W}(t)\right.\right. \\
&\left(||v_\theta(x^w_t, t, {R}_c, \mathcal{B}, \mathcal{D})-v^w_t||^2_2 - ||v_{ref}(x^w_t, t, \mathcal{R}_c, \mathcal{R}_s)-v^w_t||^2_2 \right. \\
&\left.\left.\left. - ||v_\theta(x^l_t, t, {R}_c, \mathcal{B}, \mathcal{D})-v^l_t||^2_2 + ||v_{ref}(x^l_t, t, \mathcal{R}_c, \mathcal{R}_s)-v^l_t||^2_2\right)\right)\right],
\end{split}
\end{gather}
\end{small}
where $(x^w_1, x^l_1)$ denotes the encoded latents of a collected win–lose pair, and $\mathcal{W}(t)$ is a timestep-dependent weighting factor. $v_{\text{ref}}$ refers to the reference model, which in our case is the fine-tuned text editing model. This design allows us to fully leverage the strong editing capabilities of the model learned in Stage 1.

\subsection{Transparency Glyph VAE}
To enable glyph foreground outputs in RGBA-format for flexible editing in graphic design scenarios, we design a transparency glyph VAE that supports 4-channel glyph image decoding. Inspired by prior works on layered transparent image synthesis~\cite{zhang2024transparent, pu2025art}, this module consists of a pretrained FLUX VAE encoder $\mathcal{E}_{vae}$ and a transparency VAE decoder $\mathcal{D}_{vae}$, as shown in~\cref{fig:method} (2d). The decoder is initialized from the FLUX VAE decoder and extended with additional convolution layers to produce an extra alpha channel, which is merged with the RGB channels to yield the final RGBA output. The detailed architecture is provided in Sec. 5 of the supplementary material.

During training, we sample glyphs in RGBA format $\mathcal{G}_{\text{rgba}} \in \mathbb{R}^{H \times W \times 4}$ from the \textbf{SynthGlyph Dataset}, and blend the alpha channel into the RGB channels to obtain the gray-background glyphs:
\begin{small}
\begin{equation}    
\hat{\mathcal{G}_{rgb}} = \mathcal{G}_{a}\times\mathcal{G}_{rgb}.
\end{equation}
\end{small}
We then freeze the VAE encoder and train the transparency decoder using a combination of L2 loss and LPIPS~\cite{zhang2018unreasonable} loss:
\begin{small}
\begin{gather}    \mathcal{L}_{mse}=\mathbb{E}_{\mathcal{G}_{rgba}}||\mathcal{D}_{vae}(\mathcal{E}_{vae}(\hat{\mathcal{G}_{rgb}}))-\mathcal{G}_{rgba}||^2_2, \\
\mathcal{L}_{lpips}=\mathbb{E}_{\mathcal{G}_{rgba}}||\mathcal{P}(\mathcal{D}_{vae}(\mathcal{E}_{vae}(\hat{\mathcal{G}_{rgb}})))-\mathcal{P}(\mathcal{G}_{rgba})||^2_2, \\
\mathcal{L}_{vae}=\mathcal{L}_{mse}+\lambda_{lpips}\mathcal{L}_{lpips},
\end{gather}
\end{small}
where $\mathcal{P}$ denotes the perceptual model. This training strategy ensures high-quality RGBA decoding for glyph images. When integrated with the DiT backbone, it results in a unified model capable of generating editable glyph foregrounds.

\subsection{Layout Planner}
\label{sec: layout}

In real-world scenarios of graphic design editing and generation, an appropriate layout is crucial for accurately placing generated glyph foregrounds in reasonable positions. To this end, we propose a two-stage layout planning strategy: the first stage {(coarse planning)} predicts bounding boxes for text lines, while the second stage (fine-grained planning) refines the placement by determining the position of each individual glyph within the predicted lines. This hierarchical approach enables fine-grained control over glyph positioning, which is particularly important for languages like Chinese that exhibit strong sensitivity to character-level layout.

Following prior works~\cite{seol2024posterllama, yang2024posterllava}, we adopt an MLLM as the layout planner. Conditioned on the design background and textual description, the model performs layout reasoning to infer appropriate bounding boxes for glyph placement. Inspired by previous methods, we apply SFT to the MLLM using real layout examples sampled from the \textbf{DesignText Dataset}. Specifically, we construct training samples based on predefined instruction prompts and standardized input-output formats (see Sec. 6 of the supplementary materials for more details), enabling the model to jointly learn both stages of the layout planning process.

In our experiments, we found that SFT alone was insufficient to ensure that the layout planner generates layouts that are both semantically reasonable and visually appealing. To address this limitation, we incorporated a {RL-based} approach to further optimize the MLLM. Specifically, we adopt the Group Relative Policy Optimization (GRPO) paradigm~\cite{shao2024deepseekmath} and design a set of rule-based reward functions to guide the layout generation process. Given a ground-truth layout $\mathcal{B} = \{B_1, B_2\dots B_N\}$ sampled from the dataset and a corresponding model-predicted layout $\hat{\mathcal{B}} = \{\hat{B}_1, \hat{B}_2\dots \hat{B}_N\}$, we define the following reward functions:
\begin{small}
\begin{gather}    
\mathcal{R}_{iou}= \frac1N\sum_{i=1}^{N}\frac{\mathcal{A}(\hat{B_i}\cap B_i)}{\mathcal{A}(\hat{B_i}\cup B_i)+\epsilon}, \\
\mathcal{R}_{ol}= - \frac{2}{N(N-1)}\sum_{i=1}^{N}\left(\sum_{j>i}\frac{\mathcal{A}(\hat{B_i}\cap \hat{B_j})}{\mathcal{A}(\hat{B_i}\cup \hat{B_j})+\epsilon}\right), \\
\mathcal{R}_{bl}= - \sqrt{\frac1N\sum_{i=1}^{N}(\mathcal{A}(\hat{B_i})-\mathbb{E}\mathcal{A}(\hat{\mathcal{B}}))^2}/\mathbb{E}\mathcal{A}(\hat{\mathcal{B}}), \\
\mathcal{R} = \mathcal{R}_{iou} + \lambda_{ol}\mathcal{R}_{ol} + \lambda_{bl}\mathcal{R}_{bl}, 
\end{gather}
\end{small}
where $\mathcal{A}$ denotes the area of a bounding box. The $\mathcal{R}{iou}$ term encourages accurate layout prediction by measuring the mean Intersection over Union (mIoU) between the predicted and ground-truth boxes. The $\mathcal{R}{ol}$ term penalizes overlapping predicted boxes to encourage better spatial distribution. The $\mathcal{R}_{bl}$ term penalizes size variance among the predicted boxes, promoting balanced and consistent glyph sizes across the layout.

\subsection{Inference Pipelines}

With the trained DiT model and a few off-the-shelf components, we construct a complete pipelines for stylized text editing and generation in design images, as illustrated in~\cref{fig:teaser}. For text editing, the pipeline begins by extracting the reference glyph style from a user-specified region in the original image. The DiT model then generates a new glyph foreground that matches the extracted style. To remove the original text, we apply an inpainting model~\cite{suvorov2022resolution}, obtaining a clean background. The layout planner determines the precise placement of the edited text, and the final design image is produced by merging the stylized text foreground into the background image according to the assigned position and scale.

For text generation, given a user-provided background image and textual description, the DiT model synthesizes foreground glyphs in a coherent style. The layout planner handles both line-level layout and character-level typography. The final design image is composed by merging the generated glyph foregrounds with the background image. Furthermore, by incorporating a pre-trained T2I model (e.g., FLUX), we can synthesize the background image directly from the caption, enabling a fully end-to-end text-to-design (T2D) pipeline. In summary, our system supports a broad range of applications, including high-fidelity stylized text editing and generation in design images, as well as full design image generation with accurately rendered Chinese and English text.

%% file: sec/4_experiment.tex
\section{Experiments}

\definecolor{lightblue}{rgb}{0.68, 0.85, 0.90}
\definecolor{deepblue}{rgb}{0.13, 0.55, 0.80}
\begin{table*}[htbp]
\centering
\caption{{System-level Comparison on the proposed \textbf{UTDesign-Bench}}. We highlight the \colorbox{deepblue}{best} and \colorbox{lightblue}{second-best} scores in each column.}
\label{tab: main results}
\small
\begin{tabular}{l|ccc|ccccc}
\toprule
\multirow{2}{*}{\textbf{Method}} & \multicolumn{3}{c|}{\textbf{Image Quality}} & \multicolumn{5}{c}{\textbf{Text Rendering}} \\
\cmidrule(lr){2-4} \cmidrule(lr){5-9}
 & FID$\downarrow$ & LPIPS$\downarrow$ & CLIP-Sim$\uparrow$ & Precision$\uparrow$ & Recall$\uparrow$ & F-Score$\uparrow$ & NED$\downarrow$ & Accuracy$\uparrow$ \\
\midrule
\multicolumn{9}{c}{\textbf{UTDesign-Bench-Edit}} \\
\midrule
DiffUTE~\cite{chen2023diffute} & 41.48 & 0.2676  & 0.6352 & 0.2161 & 0.1867 & 0.1967 & 0.8457 & 0.0387 \\
AnyText-Edit~\cite{tuo2024anytext} & 21.45 & \cellcolor{lightblue}0.1950 & 0.7255 & \cellcolor{lightblue}0.6274 & \cellcolor{lightblue}0.6088 & \cellcolor{lightblue}0.6049 & 0.4425 & \cellcolor{lightblue}0.3538 \\
AnyText2-Edit~\cite{tuo2024anytext2} & \cellcolor{lightblue}20.68 & 0.2042 & \cellcolor{lightblue}0.7313 & 0.5917 & 0.5753 & 0.5704 & \cellcolor{lightblue}0.4809 & 0.3029 \\
Ours & \cellcolor{deepblue}\textbf{10.81} & \cellcolor{deepblue}\textbf{0.0883} & \cellcolor{deepblue}\textbf{0.8222} & \cellcolor{deepblue}\textbf{0.9568} & \cellcolor{deepblue}\textbf{0.9482} & \cellcolor{deepblue}\textbf{0.9518} & \cellcolor{deepblue}\textbf{0.0612} & \cellcolor{deepblue}\textbf{0.8370} \\
\midrule
\multicolumn{9}{c}{\textbf{UTDesign-Bench-Gen}} \\
\midrule
Kolors & 134.2 & 0.7770 & 0.2696 & 0.0095 & 0.0106 & 0.0074 & 0.9949 & 0.0010 \\
Cogview4 & 75.46 & 0.7087 & 0.2675 & 0.3306 & 0.3898 & 0.3144 & 0.7559 & 0.0600\\
BrushYourText~\cite{zhang2024brush} & 114.4 & 0.7525 & 0.2334 & 0.7404 & 0.7451  & 0.7310 & 0.3160 & 0.4760 \\
AnyText-Gen~\cite{tuo2024anytext} & 83.57 & 0.7362 & 0.2520 & 0.5989 & 0.8232 & 0.6338 & 0.4462 & 0.2780 \\
AnyText2-Gen~\cite{tuo2024anytext2} & 90.42 & 0.7474 & 0.2509 & 0.7145 & 0.7949 & 0.7186 & 0.3541 & 0.3520 \\
Glyph-ByT5-v2~\cite{liu2024glyphv2} & 92.82 &0.6987 & 0.2465 & \cellcolor{deepblue}0.9066 &0.8971 & \cellcolor{deepblue}0.8862 & 0.2764 & \cellcolor{lightblue}0.6200 \\
PosterMaker~\cite{gao2025postermaker} & 92.10 & 0.7176 & 0.2563 & 0.7303 & 0.7382 & 0.7128 & 0.3311 & 0.4820 \\
Seedream 3.0 & \cellcolor{lightblue}72.49 & \cellcolor{deepblue}0.6903 &\cellcolor{lightblue} 0.2704 & 0.7937 & \cellcolor{deepblue}0.9603 & 0.8392 & 0.2203 & 0.4885 \\
GPT-4o & 80.93 & 0.7390 & \cellcolor{deepblue}0.2710 & 0.8457 & \cellcolor{lightblue}0.9123 & 0.8506 & \cellcolor{lightblue}0.1932 & 0.5772 \\
Ours & \cellcolor{deepblue}\textbf{72.07} & \cellcolor{lightblue}\textbf{0.6973} & \textbf{0.2609} & \cellcolor{lightblue}\textbf{0.8784} & \textbf{0.9106} & \cellcolor{lightblue}\textbf{0.8716} & \cellcolor{deepblue}\textbf{0.1590} & \cellcolor{deepblue}\textbf{0.6840} \\
\bottomrule
\end{tabular}
\end{table*}

\subsection{Benchmark and Metrics}

\sloppy
To evaluate the performance of different methods on stylized text editing and generation in graphic design, we construct a unified evaluation benchmark: \textbf{UTDesign-Bench}.
Specifically, we sample 1,000 cases from the \textbf{DesignText} test set to build two separate benchmarks: \textbf{UTDesign-Bench-Edit} and \textbf{UTDesign-Bench-Gen}, focusing on the text editing and generation capabilities, respectively.
In \textbf{UTDesign-Bench-Edit}, each case includes an original design image, the text regions and the target content. The goal is to generate a revised design image reflecting the text edits. The target text is created by randomly replacing 50\% of the original content with characters sampled from a valid character set.
In \textbf{UTDesign-Bench-Gen}, each case consists of a textual description including image caption and text to be rendered. Methods are required to generate complete design images containing the specified text, with each image containing an average of 2–3 text lines.

We evaluate models from two main perspectives: image generation quality and text rendering accuracy.
For text editing, image quality is assessed using Fréchet Inception Distance (FID)~\cite{heusel2017gans}, Learned Perceptual Image Patch Similarity (LPIPS)~\cite{zhang2018unreasonable}, which measure the visual similarity between original and edited images. Additionally, we adopt CLIP-sim to evaluate the similarity between modified and original text regions, which reflects how well the model preserves the original style.
For text generation, we also use FID and LPIPS to evaluate overall image quality. CLIP-sim is further employed to assess the alignment between the generated image and the input image caption, focusing on global semantic consistency.
To measure text rendering accuracy, we apply an off-the-shelf Optical Character Recognition (OCR) system (PP-OCRv4~\footnote{PP-OCR: \url{https://github.com/PaddlePaddle/PaddleOCR}}) to detect and recognize text in the edited or generated design images. Based on the OCR results and ground truth, we compute metrics including precision, recall, F-score, normalized edit distance (NED), and accuracy to quantitatively evaluate the correctness of text rendering. It is important to note that due to the inherent limitations of the OCR model, there exists discrepancies between the measured results and the actual performance. However, the evaluation remains fair across all methods.

\subsection{Stylized Text Editing in Design Images}

\begin{figure*}[!t]
\begin{minipage}[!t]{0.85\linewidth}
\centering
\includegraphics[width=1\textwidth]{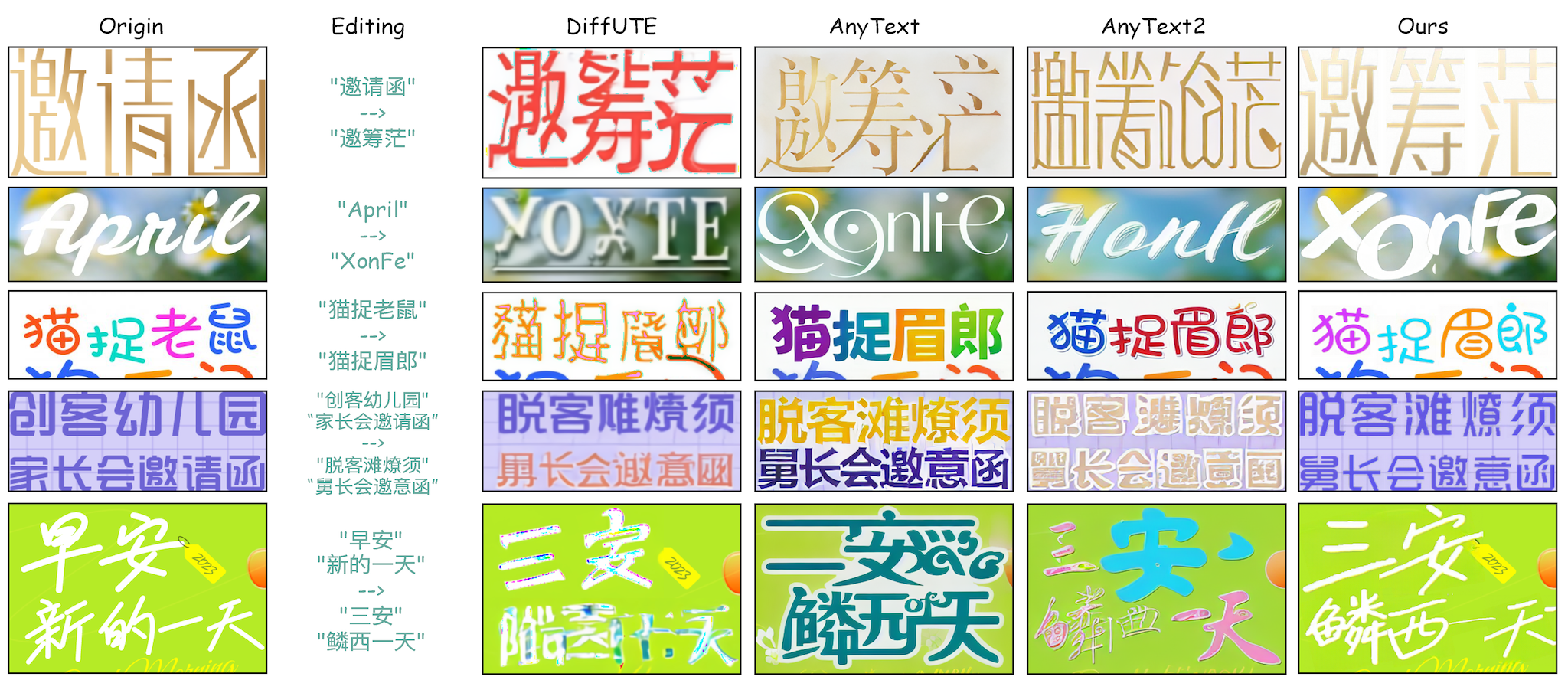}
\end{minipage}
\caption{Comparison of stylized text editing performance with strong baselines.
The first column shows original images for selected editing scenarios, with editing targets in the second column. The last three columns present results from three different methods.}
\label{fig:comp-editing}
\end{figure*}

For the editing task, we select the open-source models~\cite{chen2023diffute, tuo2024anytext, tuo2024anytext2} capable of editing both Chinese and English text in images for comparison. As shown in Table~\ref{tab: main results}, our method achieves the highest text editing accuracy, while metrics such as FID indicate that our approach not only ensures precise text edits but also maintains the consistency of the font style, outperforming baseline methods. As illustrated in~\cref{fig:comp-editing}, \textbf{UTDesign} accurately edits arbitrary text in design images while maintaining high accuracy and style consistency in both font and texture, further highlighting its superiority over existing approaches.

\subsection{Full Design Image Generation}

\begin{figure*}[!t]
\begin{minipage}[!t]{0.95\linewidth}
\centering
\includegraphics[width=1\textwidth]{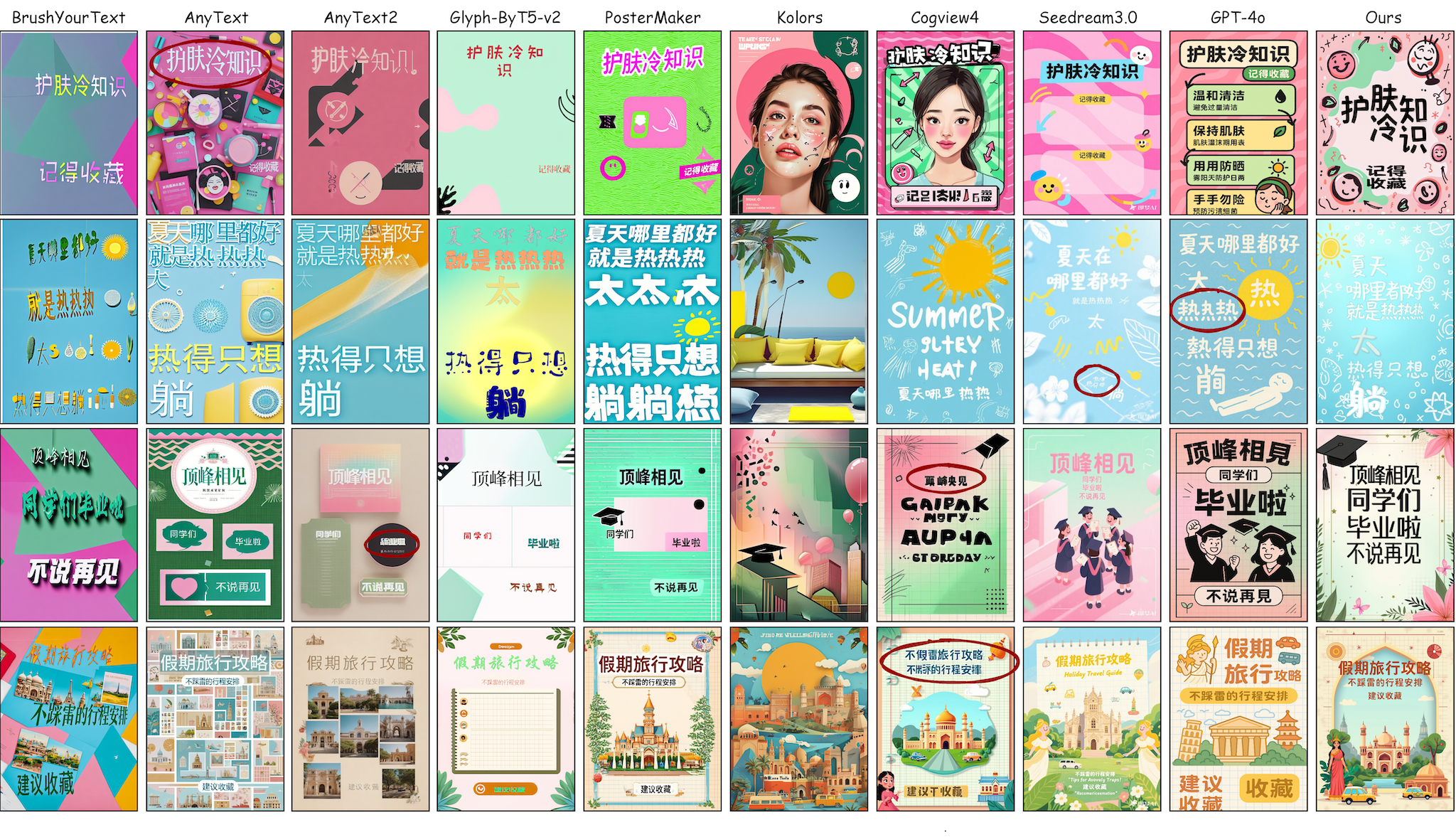}
\end{minipage}
\caption{System-level comparison with both open-source and close-source T2D models. We highlight the text rendering problems using red circles.}
\label{fig:comp-system}
\end{figure*}

\begin{figure}[!t]
\begin{minipage}[!t]{1\linewidth}
\centering
\includegraphics[width=1\textwidth]{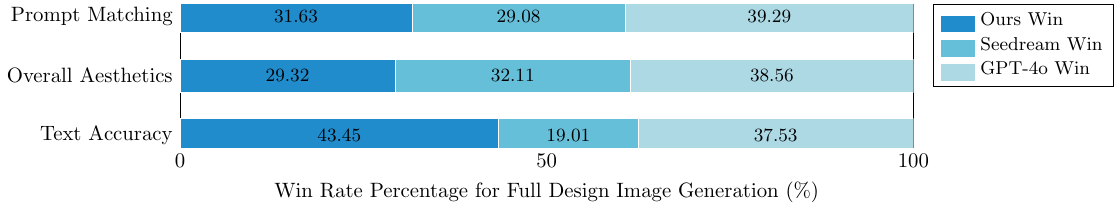}
\end{minipage}
\caption{User study comparison with proprietary commercial systems.}

\label{fig:user}
\end{figure}

For the design image generation task, we compare our method against two open-source text-to-image (T2I) models: Kolors\footnote{Kolors: \url{https://huggingface.co/Kwai-Kolors/Kolors}} and CogView4\footnote{CogView4: \url{https://github.com/THUDM/CogView4}}, both of which support Chinese and English text generation. We also include several recent bilingual text rendering methods~\cite{zhang2024brush, tuo2024anytext, tuo2024anytext2, liu2024glyphv2, gao2025postermaker} in our comparisons. For T2I models, we format the input prompt as: "Create a design image + [Description] + Text to render: + [list of texts]". For baselines that require layout inputs, we provide the ground truth layout. For Glyph-ByT5-v2 and AnyText2, we additionally sample random font and color references to meet their requirements. For PosterMaker, we provide an empty subject foreground to align the input conditions. As shown in Table~\ref{tab: main results}, our method achieves superior image quality while maintaining the highest accuracy in text rendering. It is worth noting that although open-source T2I models exhibit relatively high CLIP similarity, their overall performance remains suboptimal due to low aesthetic quality and frequent omission of visual text. We further include two proprietary commercial systems for comparison and the qualitative results of all the compared T2D models are provided in~\cref{fig:comp-system}, where our approach achieves performance on par with closed-source systems and significantly outperforms all open-source baselines. 

In practical experiments, we find that the adopted metrics do not adequately reflect human preferences, due to biases in image quality metrics such as FID and the inaccuracy of OCR systems. To address the limitations of quantitative metrics, we conduct a user study comparing our method with two proprietary approaches (Seedream 3.0 and GPT-4o) in terms of prompt matching, overall aesthetics, and text accuracy. As shown in~\cref{fig:user}, \textbf{UTDesign} demonstrates comparable overall image quality while offering a clear advantage in text rendering accuracy. More details of the user study are included in Sec. 8 of the supplementary material.

\subsection{Ablation Studies}
\begin{table}[htbp]
\centering
\caption{Ablations on the post-training of the generation model.}
\label{tab: abl-generation}
\small
\begin{tabular}{l|ccc|cc}
\toprule
\multirow{2}{*}{\textbf{Settings}} & \multicolumn{3}{c|}{\textbf{Image Quality}} & \multicolumn{2}{c}{\textbf{Text Rendering}} \\
\cmidrule(lr){2-4} \cmidrule(lr){5-6}
 & FID$\downarrow$ & LPIPS$\downarrow$ & PickScore$\uparrow$ & NED$\downarrow$ & Accuracy$\uparrow$ \\
\midrule
Aligned & 73.58 & 0.6969 & 19.83 & 0.2770 & 0.5570  \\
+ SFT &  73.12 & 0.6988 & 19.78 & 0.2624 & 0.5740 \\
+ DPO & \textbf{73.11} & \textbf{0.6968} & \textbf{20.10} & \textbf{0.2604} & \textbf{0.5930} \\
\bottomrule
\end{tabular}
\end{table}

\begin{table}[htbp]
\centering
\caption{Ablations on the layout planner.}
\label{tab: abl-layout}
\small
\begin{tabular}{l|ccc|ccc}
\toprule
\multirow{2}{*}{\textbf{Settings}} & \multicolumn{3}{c|}{\textbf{Coarse Planning}} & \multicolumn{3}{c}{\textbf{Fine-grained Planning}} \\
\cmidrule(lr){2-4} \cmidrule(lr){5-7}
 & $\mathcal{R}_{iou}\uparrow$ & $-\mathcal{R}_{ol}\downarrow$ & FID$\downarrow$ & $\mathcal{R}_{iou}\uparrow$ & $-\mathcal{R}_{ol}\downarrow$ & $-\mathcal{R}_{bl}\downarrow$ \\
\midrule
Pretrained & 0.0462 & 0.0102 & 23.77 & 0.0866 & 0.0579 & 0.5435 \\
+ SFT & 0.3637 & \textbf{0.0035} & 10.05 & 0.5883 & 0.0036 & 0.1781 \\
+ GRPO & \textbf{0.5219} & 0.0071 & \textbf{8.146} & \textbf{0.6873} & \textbf{0.0024} & \textbf{0.0912} \\
\bottomrule
\end{tabular}
\end{table}

\begin{figure*}[!t]
\begin{minipage}[!t]{0.72\linewidth}
\centering
\includegraphics[width=1\textwidth]{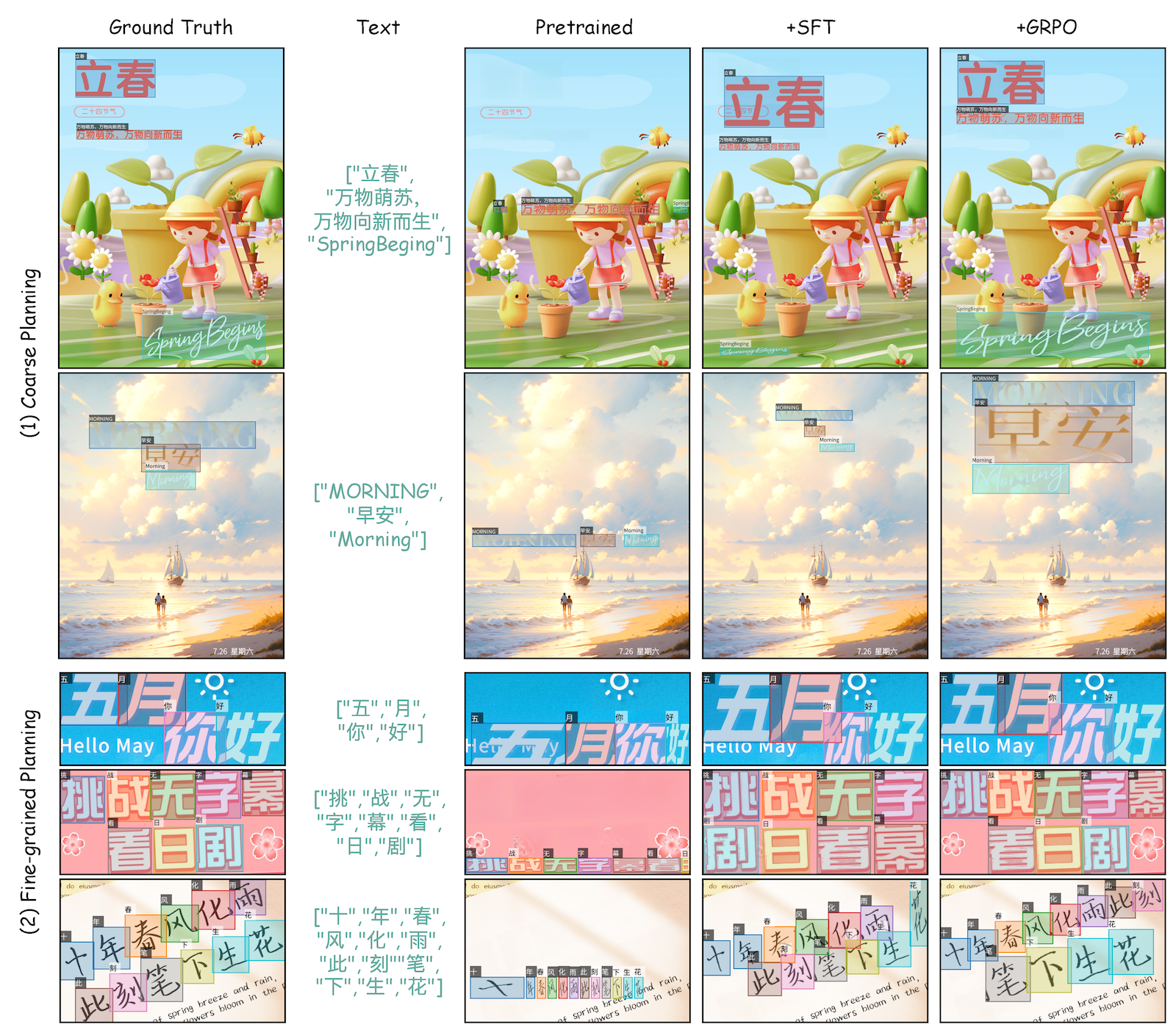}
\end{minipage}
\caption{Ablation on the training strategy of our layout planner. We show the performance of coarse planning (1) and fine-grained planning (2) at the first two rows and the last three rows, respectively. The text instances to be arranged are given at the second column.}
\label{fig:abl-layout}
\end{figure*}

\begin{figure*}[!t]
\begin{minipage}[!t]{0.76\linewidth}
\centering
\includegraphics[width=1\textwidth]{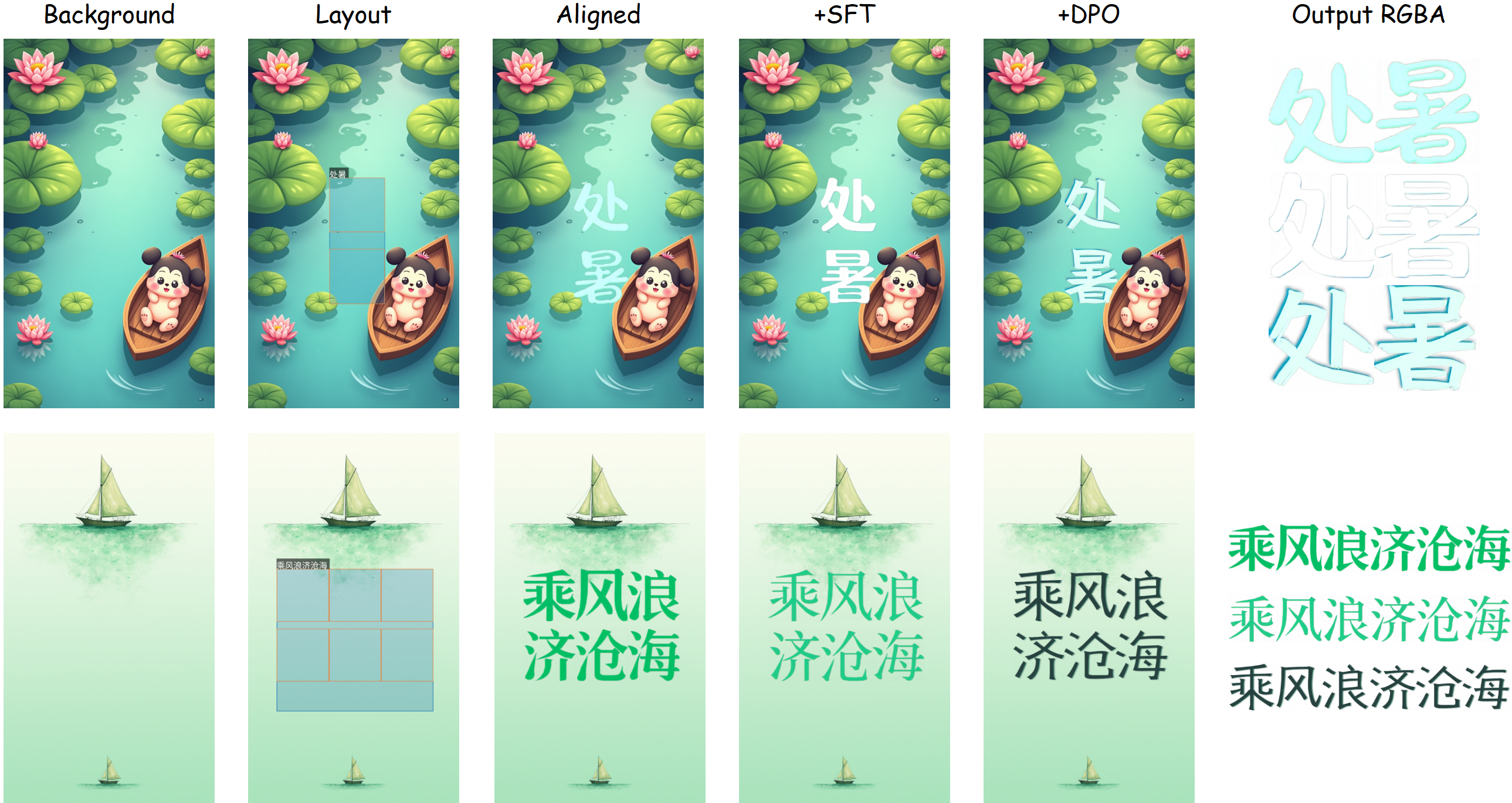}
\end{minipage}
\caption{Ablation on the post-training of our generation model. We show the background image and the corresponding layout at the first two columns. The output of our model under different training settings are listed in sequence and the output foreground glyphs in RGBA format are shown at the last column.}
\label{fig:abl-dpo}
\end{figure*}

\subsubsection{Post-training of the Generation Model}
To assess the impact of the post-training strategy, we evaluate our generation model on the \textbf{UTDesign-Bench-Gen} under three progressively enhanced configurations:
(1) \textbf{Aligned}: The base model incorporates the multi-modal encoder from Stage 2, following feature alignment, and connects it directly to the backbone network.
(2) \textbf{+SFT}: Building on (1), the model is further fine-tuned using LoRA on a high-quality subset of the \textbf{DesignText Dataset}.
(3) \textbf{+DPO}: Extending (2), the model is additionally trained with DPO using sampled win-lose pairs.

For fair comparison, we prepare the background images and the corresponding layouts in advance and run inference under same conditions. Beyond standard benchmark metrics that evaluate image generation quality and text rendering accuracy, we also incorporate PickScore~\cite{kirstain2023pick} to measure the aesthetic quality of the generated outputs.
As shown in~\cref{tab: abl-generation} and ~\cref{fig:abl-dpo}, SFT consistently improves performance, while DPO further enhances both the aesthetic appeal and diversity of the generated design images.

\subsubsection{Layout Planner}
To validate the effectiveness of our training strategy, we evaluate the performance of our layout planner under three progressively refined configurations:
(1) \textbf{Pretrained}: The pretrained MLLM is directly used for layout planning based on the given instruction.
(2) \textbf{+SFT}: Building on (1), the MLLM is fine-tuned on the \textbf{DesignText Dataset}.
(3) \textbf{+GRPO}: Extending (2), the MLLM is further optimized using GRPO paradigm.

We evaluate the model in both coarse and fine-grained layout planning on a subset of the \textbf{DesignText Dataset}.
For coarse planning, we directly assess the layout quality using mIoU and overlay reward scores {($\mathcal{R}_{iou}$, $\mathcal{R}_{ol}$)} as described in~\cref{sec: layout}. The glyph foregrounds are then rendered into the predicted boxes to composite a full design, and FID is computed to assess the overall image quality.
For fine-grained planning, we additionally introduce a balance reward score {$\mathcal{R}_{bl}$} to measure the evenness of glyph box distribution.

As shown in~\cref{tab: abl-layout}, the pretrained MLLM alone is insufficient for high-quality layout planning. Fine-tuning with SFT significantly improves task adaptation, while GRPO further enhances the quality and coherence of the predicted layouts.
Please refer to~\cref{fig:abl-layout} for visualizations of layout planning performance at both the coarse and fine-grained planning stages.

%% file: sec/5_conclusion.tex
\section{Limitation}
Although our method can produce high-quality and high-fidelity glyph foregrounds, it still has certain limitations.
First, due to the use of DiT architecture and our implementation based on sequence concatenation, the inference time increases nonlinearly with the number of characters rendered.
In addition, the stylistic diversity of the generated glyphs remains limited, primarily due to constraints in the training dataset. This issue could potentially be alleviated by leveraging larger and higher-quality datasets in the future. Some of our failure cases can be found in the supplementary material.

\section{Conclusion}
In this paper, we proposed a unified AI-assisted framework for high-precision text editing and generation in graphic design images, supporting both English and Chinese scripts. Mainly by adopting a DiT-based model with a transparency glyph VAE, our system generates stylized text in an editable RGBA format that preserves the style of the reference glyphs, achieving state-of-the-art results in both text accuracy and stylistic consistency. Moreover, we introduced two specialized datasets and extended the model into a text generation framework conditioned on background image and text description using a multi-modal condition encoder. By integrating T2I models and an MLLM-based layout planner, we developed an end-to-end T2D pipeline that converts user intentions into complete graphic designs. The proposed framework holds strong potential for real-world applications such as personalized poster editing and glyph design, automated advertising creation, etc.

\begin{acks}
This work was supported by National Natural Science Foundation of China (Grant No.: 62372015), Leading Projects in Key Research Fields of Language Funded by the National Language Commission, Center For Chinese Font Design and Research, Key Laboratory of Intelligent Press Media Technology, and State Key Laboratory of General Artificial Intelligence.
\end{acks}

%% file: appendix.tex
\begin{appendix}

\section{System Implementation Details}
For \textbf{Model Architecture}, the DiT model consists of 16 Fusion DiT Blocks followed by 8 Single DiT Blocks. The glyph content encoder is based on DINOv2-Large, while the glyph style encoder adopts CLIP-ViT-Large-Patch14 as its backbone. The MLLM backbone used in the multi-modal condition encoder and the layout planner are  Qwen-2.5-VL-3B-Instruct and Qwen-2.5-VL-7B-Instruct, respectively. The perceiver resampler employs query vectors with a total length of 8×256. For \textbf{Training Details}, we fix the glyph generation resolution at 256×256. In Stage 1, we train the model for 120k steps with a batch size of 32, with 60k steps on gray-scale stylized fonts and another 60k steps on stylized fonts enhanced with color and texture. In Stage 2, we align the multi-modal condition encoder with 8k steps of training using a batch size of 16. In Stage 3, we fine-tune the DiT model using LoRA with a batch size of 16, including 4k steps of SFT and 4k steps of DPO. {The DPO training dynamics can be seen in~\cref{fig:curves}}. For the transparency glyph VAE, we train at a resolution of 1024×1024 for 73k steps with a batch size of 128. For \textbf{Hyperparameters}, we set $\lambda_{lpips}=0.1$, $\lambda_{ol}=0.5$, and $\lambda_{bl}=0.5$. All optimization is performed using the Prodigy optimizer~\cite{mishchenko2024prodigy} with a learning rate of 1.0. Experiments are conducted on 4x NVIDIA A100 80GB GPUs. For inference, we utilize the overshooting sampling algorithm~\cite{hu2024amo} with a cfg scale of 3.5.

\section{More Details of the DesignText Dataset}

\begin{figure*}[!t]
    \centering
    \begin{minipage}[c]{0.48\textwidth}
        \centering
        \includegraphics[width=\textwidth]{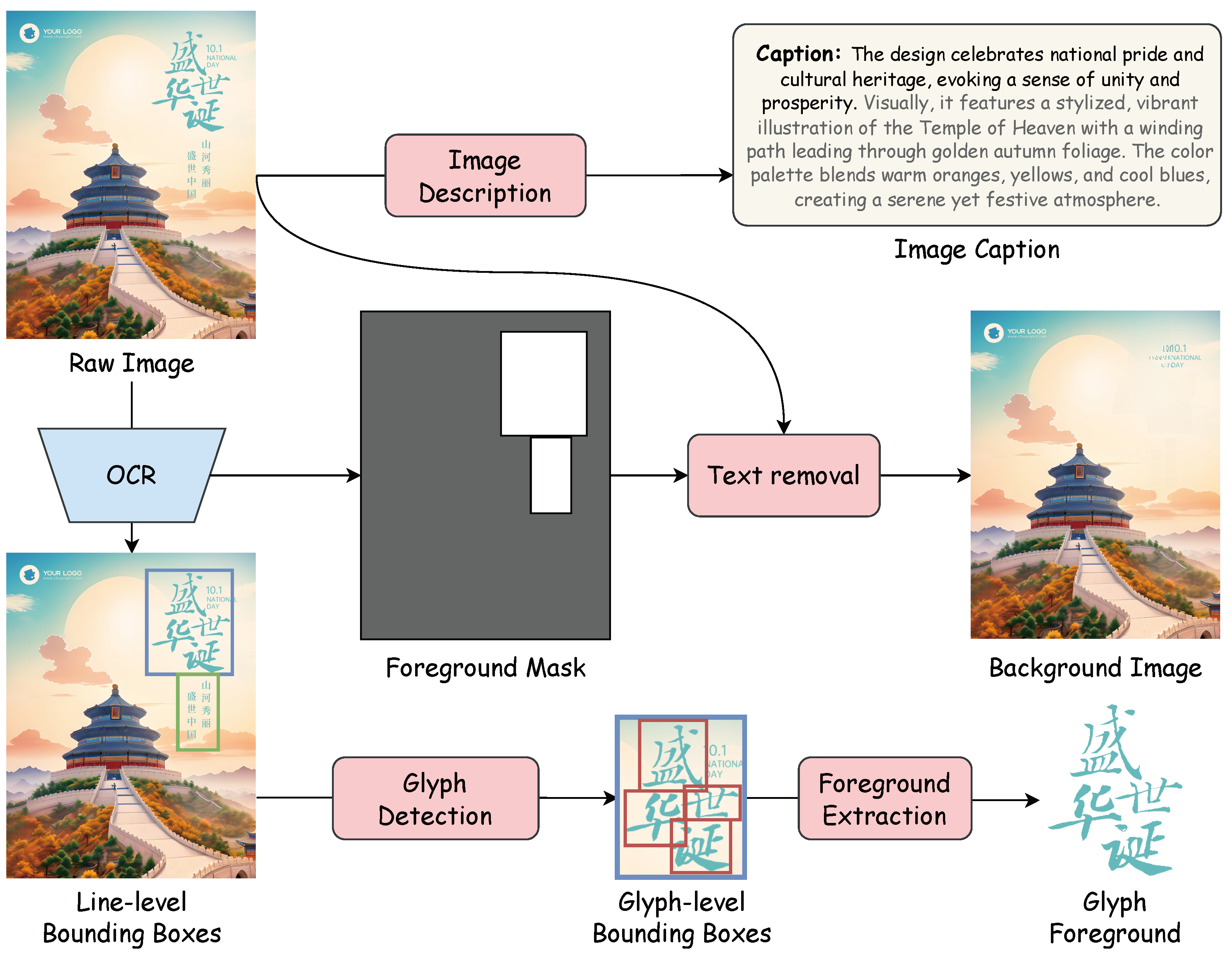}
        \caption{Building pipeline of the DesignText Dataset.}
        \label{fig:preprocess}
    \end{minipage}
    \begin{minipage}[c]{0.45\textwidth}
        \centering
        \includegraphics[width=\textwidth]{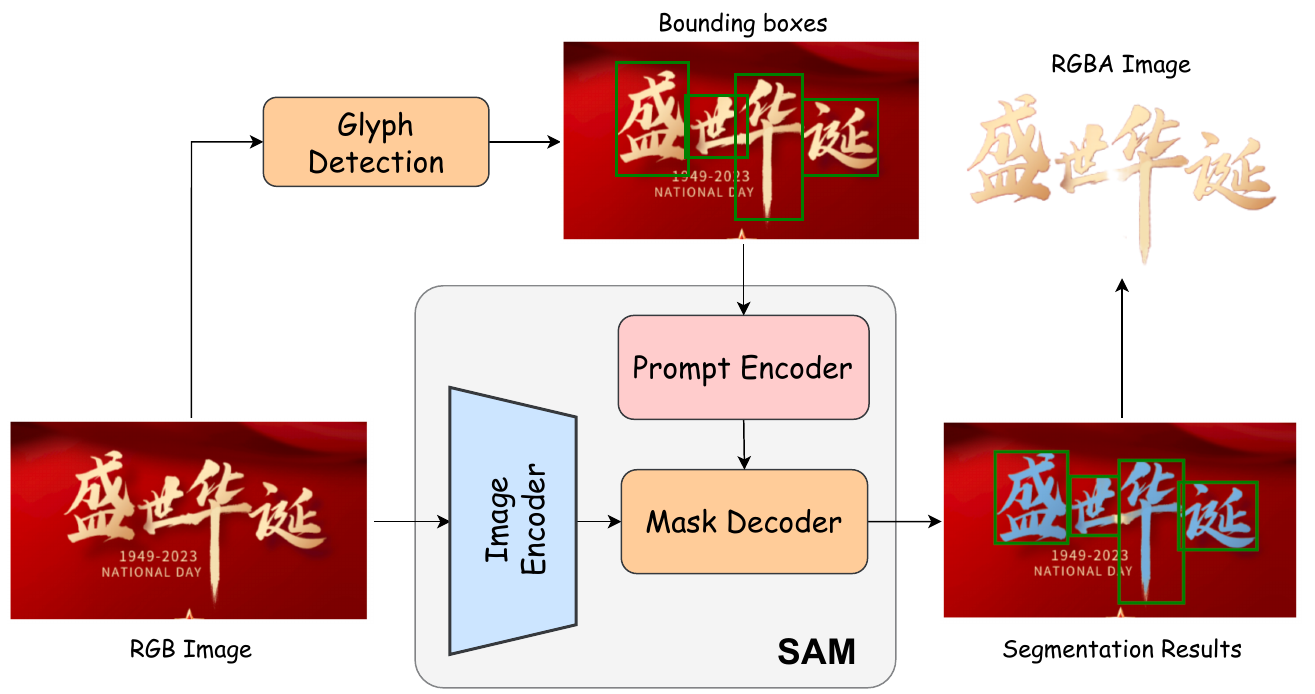}
        \caption{Glyph foreground extraction.}
        \label{fig:sam}
    \end{minipage}
    \label{fig:data-pipeline}
\end{figure*}

\begin{figure*}[!t]
\begin{minipage}[!t]{0.9\linewidth}
\centering
\includegraphics[width=1\textwidth]{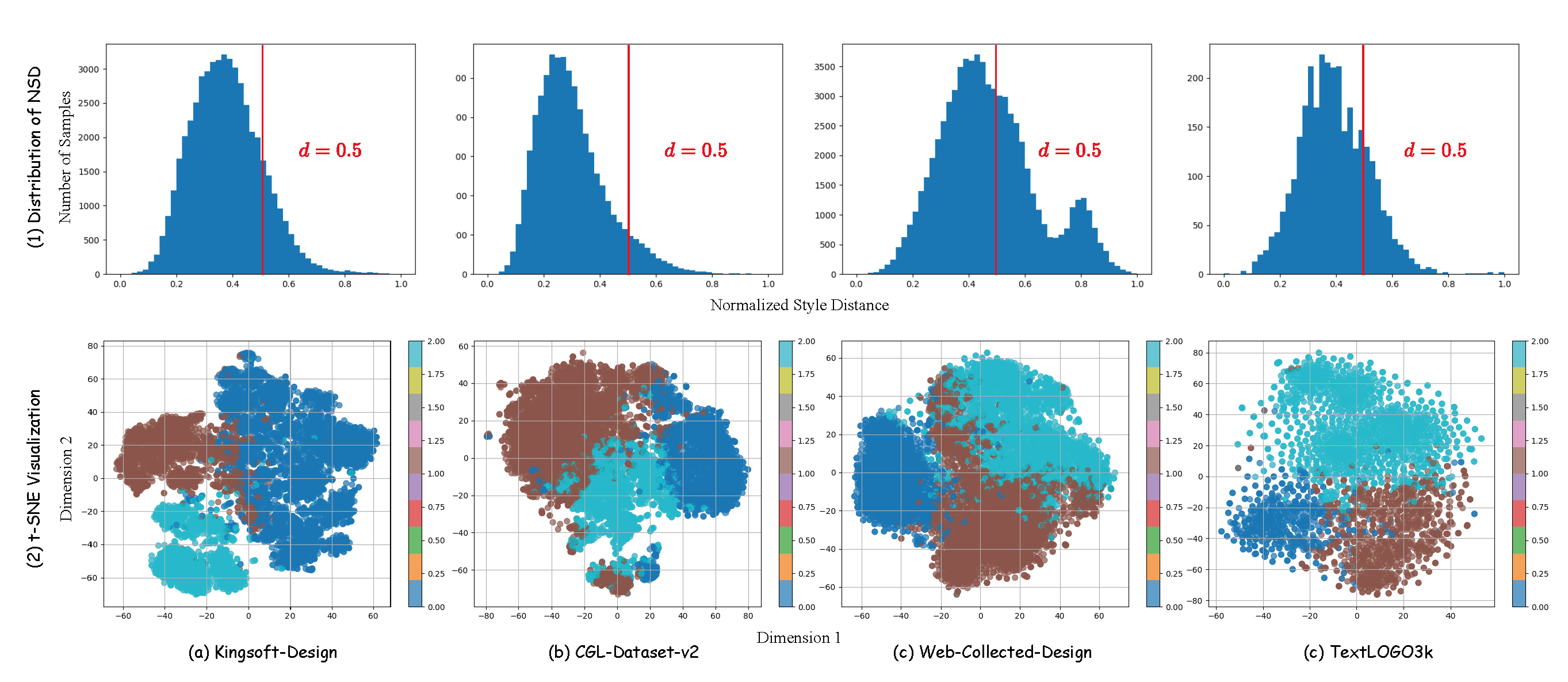}
\end{minipage}
\caption{Data distribution of the DesignText Dataset. The first row shows the frequency histogram of the normalized style distance for each data source. The second row illustrates the corresponding t-SNE visualization of the style features.}
\label{fig:data-filtering}
\end{figure*}

\subsection{Data Sources}
We collect data for the \textbf{DesignText Dataset} from these different data sources:
\sloppy
\begin{enumerate} 
    \item \textbf{Kingsoft-Design}: An internal dataset containing 28.1k graphic design samples, covering horizontal and vertical posters, banners, letters, advertisements, and more. This dataset has been meticulously annotated with character-level content and bounding box information. 
    \item \textbf{CGL-dataset-v2}~\cite{li2023relation}: A publicly available dataset with 60.5k samples, primarily composed of advertising posters, providing text-line level annotations. 
    \item \textbf{Web-Collected-Design}: A dataset of 23.8k design images collected from public websites, spanning a wide range of themes and styles. 
    \item \textbf{TextLOGO3k}~\cite{wang2022aesthetic}: A finely annotated Chinese poster dataset comprising 3k high-quality movie posters with detailed foreground text annotations.
\end{enumerate}

\subsection{Building Pipeline of the \textbf{DesignText Dataset}}
To handle data sources with varying levels of annotation detail and to construct a unified, fine-grained annotated dataset for design images, we develop a comprehensive preprocessing pipeline that extracts titles, backgrounds, glyph bounding boxes, and foregrounds from raw RGB design images.
As illustrated in~\cref{fig:preprocess}, we first apply an off-the-shelf OCR system (i.e., PP-OCRv4) to the raw RGB images to detect text lines and obtain line-level bounding boxes. We then perform glyph-level detection on the cropped text lines using a fine-tuned object detection model (i.e., YOLOv11n) to extract glyph-level bounding boxes. Finally, a well-designed glyph foreground extraction method is applied to obtain glyph foreground images in RGBA format.
To obtain clean design backgrounds, we first generate a binary mask based on the OCR-detected text regions. We then use a Fourier-based inpainting method~\cite{suvorov2022resolution} to remove glyph areas and recover the background.
For image captioning, we utilize a multi-modal large language model (i.e., GPT-4o) to describe each design image in the dataset, summarizing its theme and visual characteristics. The instruction used is as follows:
\begin{tcolorbox}[
    enhanced,
    title=Image Captioning Instruction,
    fonttitle=\bfseries,
    coltitle=black,
    colframe=blue!30!black,
    colback=blue!5!white!70,
    colbacktitle=blue!20!white,
    sharp corners=south,
    boxrule=0.7pt,
    drop shadow southeast,
    width=\linewidth,
    arc=2mm,
    left=2mm,
    right=2mm,
    top=1mm,
    bottom=1mm,
    before skip=10pt,
    after skip=10pt
]
\textbf{Instruction}: <image>Please first describe the main motivation (theme) of the design image and then describe the visual appearance, focusing on color, texture, patterns, and overall style. \
        Keep the description concise and avoid describing any text present in the image. The description should be in English and not exceed 100 words. Don't use line breaks. \\
\textbf{Output}:  The design image is about... Visually, it features...
\end{tcolorbox}
It is worth noting that this pipeline is designed with procedural flexibility: whenever available, existing annotations from the data source (e.g., text line bounding boxes) can be leveraged to replace automated components of the pipeline (e.g., OCR detection), thereby yielding more accurate annotation results.

\subsection{Glyph Foreground Extraction}
To accurately extract glyph foregrounds from design images, we propose a foreground extraction method based on a glyph detection model (i.e., YOLO~\cite{khanam2024yolov11}) and an instance segmentation model (i.e., SAM~\cite{kirillov2023segment}), as illustrated in~\cref{fig:sam}.
Given a cropped text region $\mathcal{R}_{rgb}$, we first use the glyph detection model to extract character-level bounding boxes $\mathcal{B} = \{B_1, B_2, \dots, B_n\}$. These boxes are then used as prompts for the SAM model to perform foreground segmentation. Finally, the predicted foreground masks are merged with the RGB channels as the alpha channel to produce an output in RGBA format.

\begin{figure*}[!t]
\begin{minipage}[!t]{0.75\linewidth}
\centering
\includegraphics[width=1\textwidth]{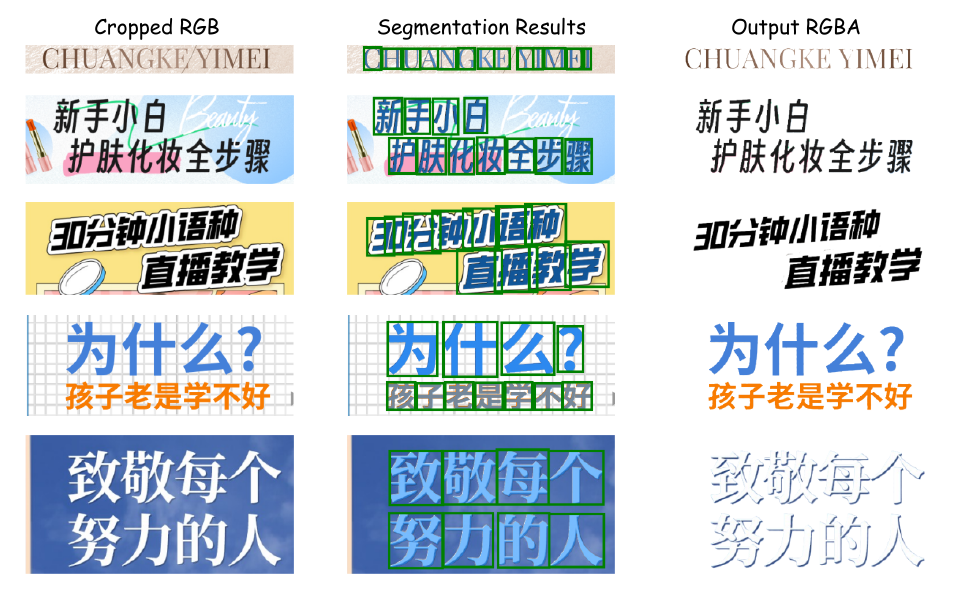}
\end{minipage}
\caption{Showcase the performance of our glyph foreground extraction method.}
\label{fig:sam-output}
\end{figure*}

Since the pretrained SAM model only outputs binary masks and generalizes poorly to glyph segmentation tasks, we introduce several modifications. First, we remove the model’s final binarization step and apply clipping to constrain the output within a continuous range, resulting in $\hat{\mathcal{R}}_{\alpha} = \mathcal{SAM}(\mathcal{R}_{rgb}, \mathcal{B}) \in [0,1]^{h \times w}$.
We then fine-tune the model in a supervised manner on a synthetic dataset. Given the ground-truth alpha channel $\mathcal{R}_{\alpha}$, the model is trained using the following loss function:
\begin{small}
\begin{gather}
\mathcal{L}_{mse} = \mathbb{E}_{\mathcal{R}_{\alpha}} ||\hat{\mathcal{R}_{\alpha}} - \mathcal{R}_{\alpha}||^2_2, \\
p_t(x,y) = \begin{cases} \hat{\mathcal{R}_{\alpha}}(x,y), & \text{if } \mathcal{R}_{\alpha}(x,y) > 0 \\ 1-\hat{\mathcal{R}_{\alpha}}(x,y), & \text{otherwise} \end{cases} \\
\mathcal{L}_{focal} = - \mathbb{E}_{\mathcal{R}_{\alpha}} \alpha (1 - p_t)^{\gamma}\log(p_t), \\
\mathcal{L}_{dice} = 1 -\mathbb{E}_{\mathcal{R}_{\alpha}} \frac{2 ||\hat{\mathcal{R}_{\alpha}}  \mathcal{R}_{\alpha}||_1}{||\hat{\mathcal{R}_{\alpha}}||_1 + ||\mathcal{R}_{\alpha}||_1}, \\
\mathcal{L} = \mathcal{L}_{mse} + \lambda_{focal} \mathcal{L}_{focal} + \lambda_{dice} \mathcal{L}_{dice},
\end{gather}
\end{small}
where $\mathcal{L}_{mse}$ measures the pixel-wise difference between the predicted and ground-truth alpha masks, $\mathcal{L}_{focal}$ evaluates the accuracy of foreground-background classification, and $\mathcal{L}_{dice}$ assesses the overlap between the predicted and ground-truth foreground regions by computing mIoU. 
After training, the SAM model adapts well to the task of glyph foreground extraction and produces high-quality foreground outputs, as illustrated in ~\cref{fig:sam-output}.

\subsection{Data Filtering of the DesignText Dataset}
\begin{figure}[H]
\begin{minipage}[H]{0.8\linewidth}
\centering
\includegraphics[width=1\textwidth]{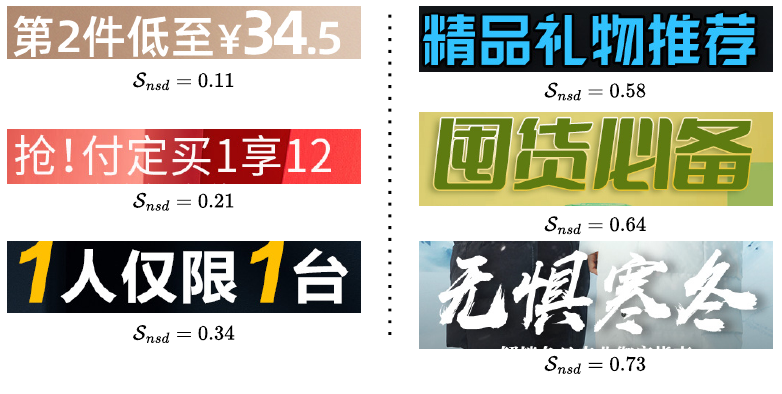}
\end{minipage}
\caption{Illustrating the data samples with different NSD score.}
\label{fig:nsd}
\end{figure}

During our experiments, we observed that directly training the generation model on the full \textbf{DesignText Dataset} yielded suboptimal results. This is likely due to the presence of a large number of trivial samples (design images using fonts close to standard typefaces with plain colors and textures) which led to limited stylistic diversity and lower aesthetic quality in the generated glyphs.
To address this issue, we propose a data filtering method based on the Normalized Style Distance (NSD) to select glyph samples with higher stylistic diversity and aesthetic quality.
Specifically, we first use the style encoder $\mathcal{S}$ trained in Stage 1 to extract normalized style features $s_i = \mathcal{S}(\mathcal{G}_i)$ for each glyph sample $\mathcal{G}_i$ in the dataset. We then perform K-Means clustering on all style features to obtain $k$ cluster centers $\{c_1, c_2\dots c_k\}$. A visualization of the clustering result is shown in~\cref{fig:data-filtering} (2).
For each glyph sample $\mathcal{G}_i$, its NSD value is defined as the Euclidean distance from its style feature $s_i$ to the nearest cluster center in the feature space:
\begin{small}
\begin{equation}
\mathcal{S}_{nsd}(\mathcal{G}_i) = \min_{1 \leq j \leq k} \sqrt{\|s_i - c_j\|^2_2} \in [0, 1],
\end{equation}
\end{small}
The NSD distribution across different data sources is shown in~\cref{fig:data-filtering} (1). Based on a predefined threshold $d$, we filter out samples with $s_i > d$, thereby selecting glyphs that exhibit greater stylistic variation and higher aesthetic quality. The threshold $d$ can be adjusted to control the quality level of the selected data.
As shown in~\cref{fig:nsd}, the NSD metric effectively reflects the aesthetic appeal and stylistic richness of glyph samples.

\section{3D-RoPE in the \textbf{UTDesign}-DiT}

\begin{figure}[H]
\begin{minipage}[H]{1.0\linewidth}
\centering
\includegraphics[width=1\textwidth]{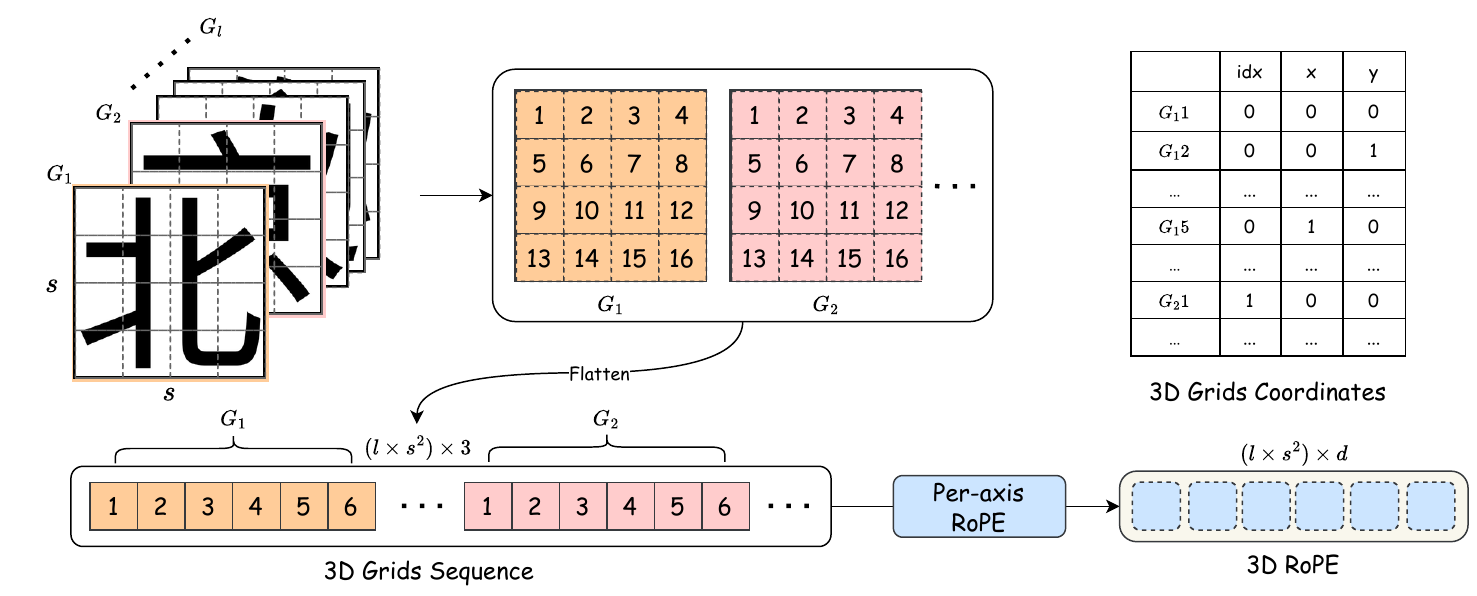}
\end{minipage}
\caption{3D-RoPE implementation in the proposed \textbf{UTDesign}-DiT.}
\label{fig:3d-rope}
\end{figure}


To support an arbitrary number of glyph outputs, we design a customized 3D Rotary Position Embedding (3D-RoPE) for encoding sequence positional information within the \textbf{UTDesign}-DiT.
As illustrated in ~\cref{fig:3d-rope}, given a glyph sequence $\mathcal{G} = \{G_1, G_2, \dots, G_l\}$ of length $l$, and assuming each glyph latent has a spatial resolution of $s \times s$, we construct a grid of the same size. Each grid cell $G_{ij}$ is assigned a 3D coordinate $(\text{idx}, x, y)$, where $\text{idx} = i - 1$ distinguishes which glyph the current grid point belongs to, and $(x, y)$ represents its 2D spatial position.
By flattening the grid sequence and concatenating them in glyph order, we obtain a 3D grid sequence $\mathcal{G} = \{G_1, G_2\dots G_k\} \in \mathbb{N}^{k \times 3}$, where $k = l \times s^2$.
Finally, we apply RoPE calculation independently along each of the three axes and concatenate the resulting embeddings to obtain the final 3D RoPE:
\begin{small}
\begin{equation}
\mathcal{R} = \{R_1, R_2\dots R_k\} \in \mathbb{R}^{k \times d},
\end{equation}
\end{small}

\section{Implementation of the Perceiver Resampler}

\begin{figure}[H]
\begin{minipage}[H]{0.8\linewidth}
\centering
\includegraphics[width=1\textwidth]{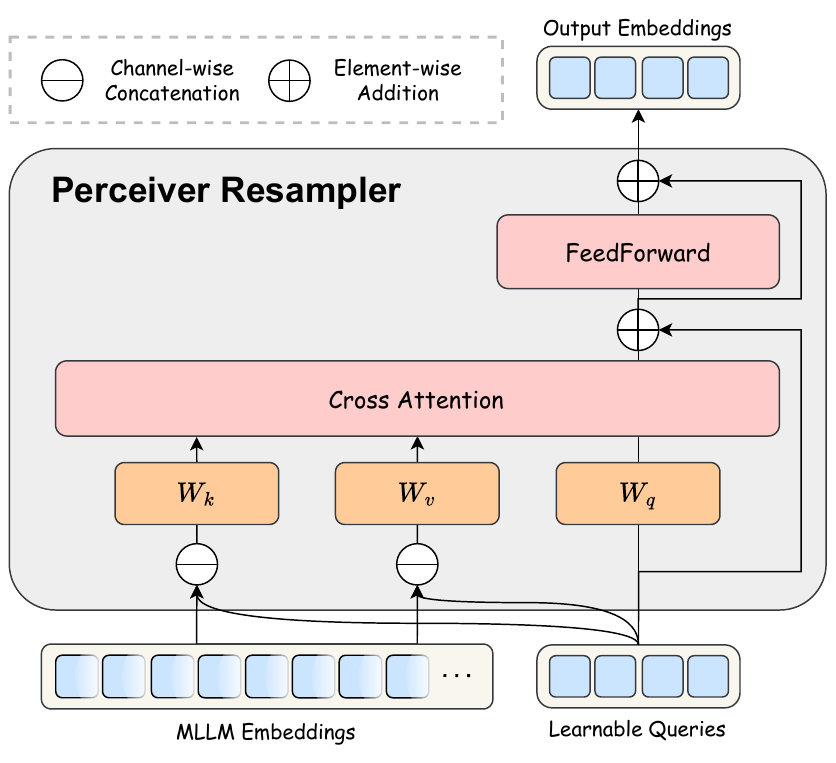}
\end{minipage}
\caption{Model structure of the perceiver resampler.}
\label{fig:perceiver}
\end{figure}

To efficiently extract fixed-length embeddings from the output of the MLLM-based multi-modal condition encoder for feature alignment, we adopt the Perceiver Resampler~\cite{alayrac2022flamingo} architecture. As illustrated in~\cref{fig:perceiver}, given a set of learnable queries $\mathcal{Q} = \{q_1, q_2, \dots, q_l\}$ of length $l$, we first concatenate them with the MLLM embeddings $\mathcal{M} = \{m_1, m_2, \dots, m_k\}$ and apply key-value projection. The resulting representations are then passed through a sequence of cross-attention and feedforward layers to produce the final output embeddings $\mathcal{O} = \{o_1, o_2, \dots, o_l\}$. The formal expression is given by:

\begin{small}
\begin{gather}
\mathcal{Q}' = \text{cross-attention}(\mathcal{Q}, [\mathcal{Q}, \mathcal{M}]) + \mathcal{Q}, \\
\mathcal{O} = \text{feed-forward}(\mathcal{Q}') + \mathcal{Q}',
\end{gather}
\end{small}

\section{Implementation of the Transparency VAE Decoder}

\begin{figure}[H]
\begin{minipage}[H]{1.0\linewidth}
\centering
\includegraphics[width=1\textwidth]{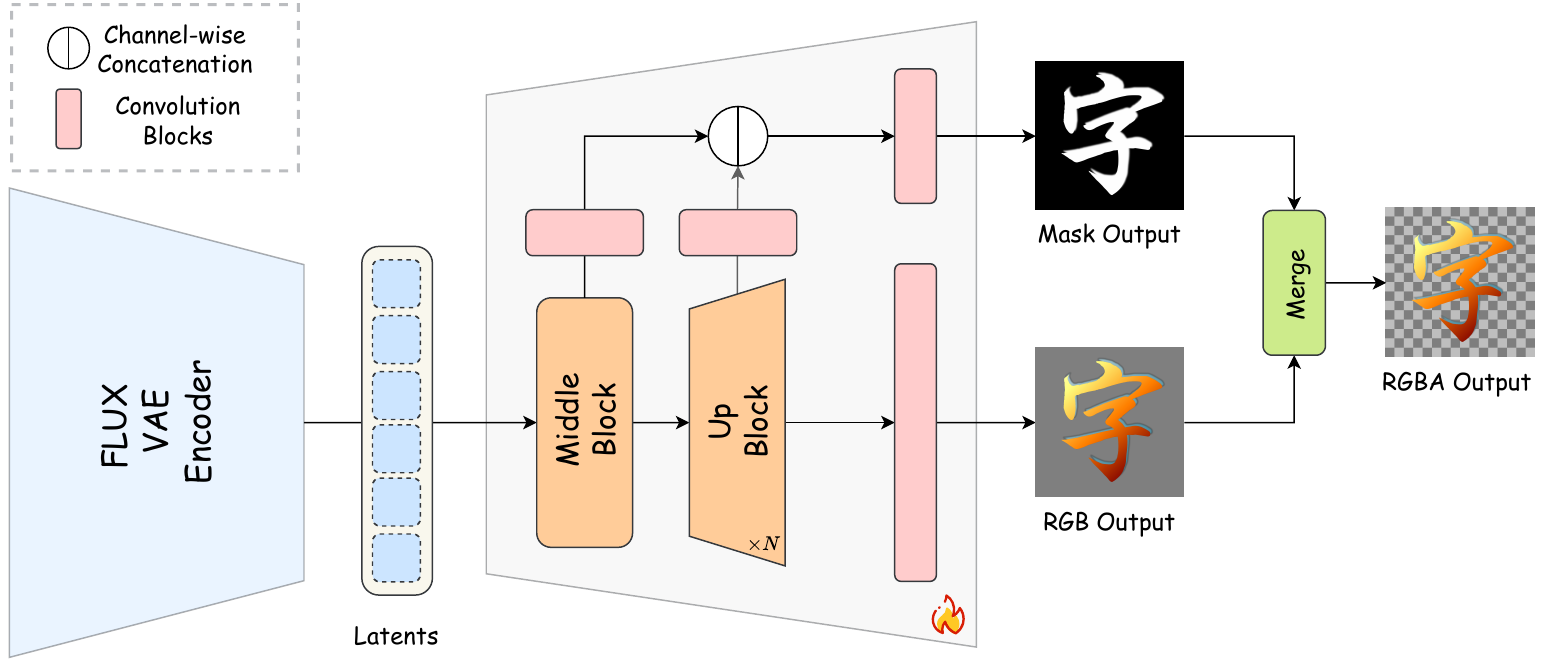}
\end{minipage}
\caption{Model structure of the proposed transparency VAE decoder.}
\label{fig:trans-vae}
\end{figure}

In \textbf{UTDesign}, we adopt the FLUX VAE encoder to encode latent representations. To reconstruct RGBA glyph images from these latents, we introduce several modifications to the original FLUX VAE decoder, as shown in~\cref{fig:trans-vae}. Specifically, while retaining the backbone architecture, we insert additional convolutional layers after each middle block and up block to extract residual features. These residuals are concatenated and passed through a convolutional layer to predict a single-channel alpha mask. The predicted alpha channel is then fused with the RGB output from the backbone to produce the final 4-channel RGBA image.
In practice, we optimize the decoder using an MSE loss to minimize the pixel-wise difference between the predicted and ground truth RGBA images, and incorporate an LPIPS loss to enhance perceptual quality. The decoder is trained at a high resolution of 1024px, and we empirically observe that the high-resolution VAE generalizes well to lower-resolution tasks (e.g., 256px), producing high-quality glyph reconstructions.

\section{More Details of the Layout Planner}
\subsection{Related Works on Layout Planning for Design Image Generation}
Layout planning is regarded as a crucial task in design image generation, as it directly impacts both the effectiveness of information delivery and the overall aesthetic appeal.
To reduce manual labor, various methods have been proposed to automate layout design.
Some approaches~\cite{hsu2023posterlayout, li2024cgb, li2023relation} attempt to learn layout generators using GANs or diffusion models.
More recent methods~\cite{lin2023layoutprompter, tanglayoutnuwa} leverage large language models (LLMs) to perform text-to-layout generation by defining structured output formats for LLMs.
Other works~\cite{seol2024posterllama, yang2024posterllava} incorporate visual conditions (e.g., background images) and enhance layout planning capabilities with the help of multi-modal large language models (MLLMs), further improving results through preference alignment~\cite{patnaik2025aesthetiq}.
Additionally, a few methods~\cite{wang2022aesthetic, he2024gldesigner} focus on glyph-level layout planning within designated regions, enabling more precise control over artistic text layout—particularly for Chinese characters.
In this work, we propose a two-stage layout planning approach based on MLLMs to achieve both region-level text layout design and fine-grained glyph-level planning.
\subsection{The Input/Output Definition of the Proposed Layout Planner}
Following previous works, we transfer the MLLM into a layout planner by getting responses from the model using well-designed instruction prompts. To ensure machine-readability and downstream compatibility, we employ a structured JSON schema to standardize the model outputs based on the structured output paradigm. The predefined instruction prompts and standardized output formats are as follows:
\begin{tcolorbox}[
    enhanced,
    title=Layout Planning Input/Output Format,
    fonttitle=\bfseries,
    coltitle=black,
    colframe=blue!30!black,
    colback=blue!5!white!70,
    colbacktitle=blue!20!white,
    sharp corners=south,
    boxrule=0.7pt,
    drop shadow southeast,
    width=\linewidth,
    arc=2mm,
    left=2mm,
    right=2mm,
    top=1mm,
    bottom=1mm,
    before skip=10pt,
    after skip=10pt
]
\textbf{Instruction}: <image>Please help me design a layout to place \{$l$\} foreground text items over the background of original size w=\{$w$\}, h=\{$h$\}. \{caption\} The foreground text items are \{labels\}. Place the items carefully to avoid unbalance, overlap, and out-of-bounds. The layout should contain all the text items in given order, in which each item has a bounding box described as [left, top, right, bottom] (all the values are integer numbers). Return the result by filling in the initial JSON file while keeping the label of items unchanged and do not return any extra explanation. The initial JSON is defined as: \{JSON template\}. \\
\textbf{Output}:  [
    \{
        "label": "hello",
        "bbox": [100, 200, 300, 400]
    \},
    ...
]
\end{tcolorbox}

\subsection{{Training Dynamics}}

\begin{figure}[H]
\begin{minipage}[H]{1.0\linewidth}
\centering
\includegraphics[width=1\textwidth]{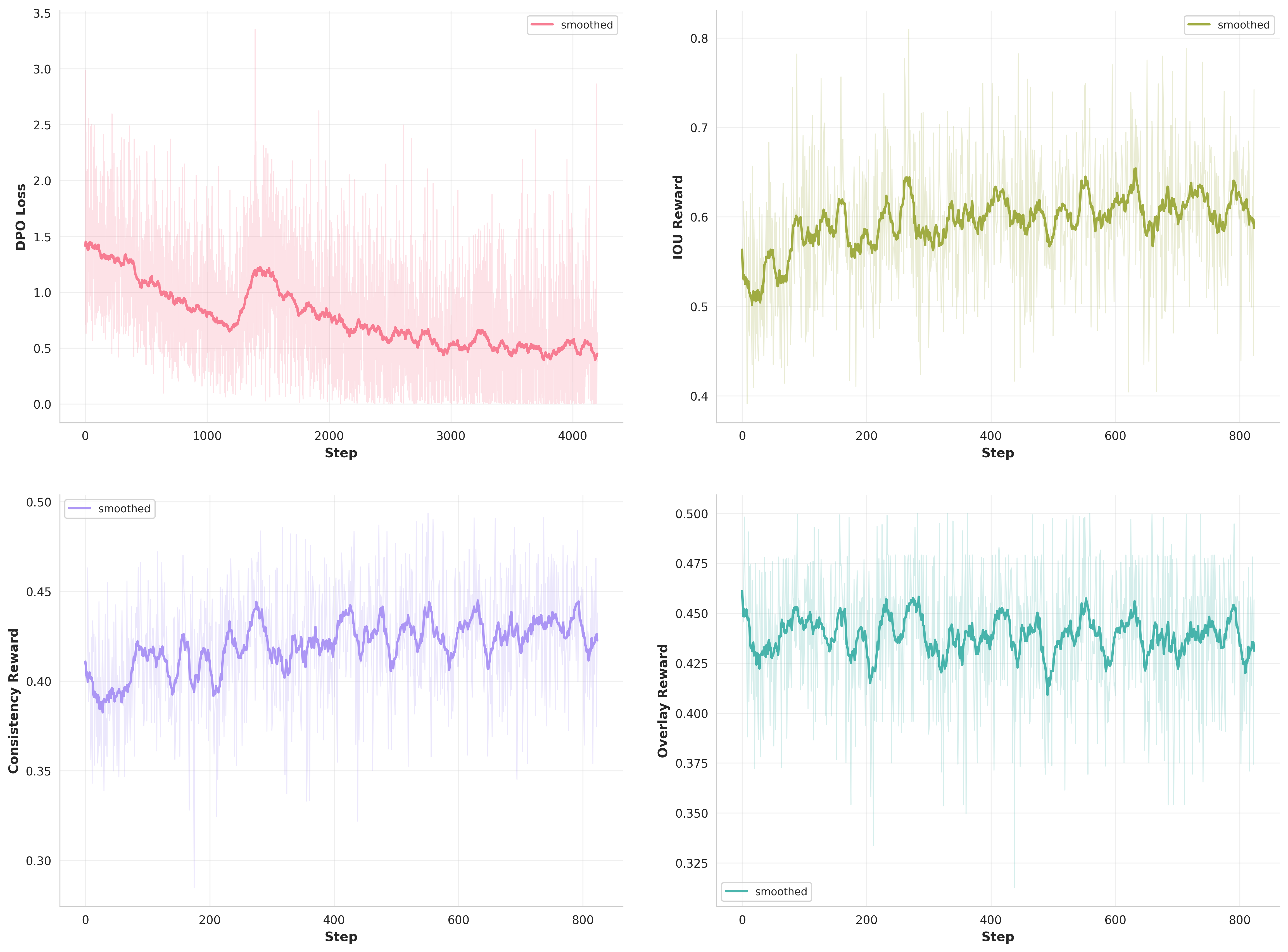}
\end{minipage}
\caption{Visualization of the RL training dynamics.}
\label{fig:curves}
\end{figure}

As shown in~\cref{fig:curves}, we visualize the training dynamics of the layout planner during the reinforcement learning process. It can be seen that as the training progresses, the model's reward score has been effectively improved.

\section{More Qualitative Results}
\begin{figure*}[!t]
\begin{minipage}[!t]{0.8\linewidth}
\centering
\includegraphics[width=1\textwidth]{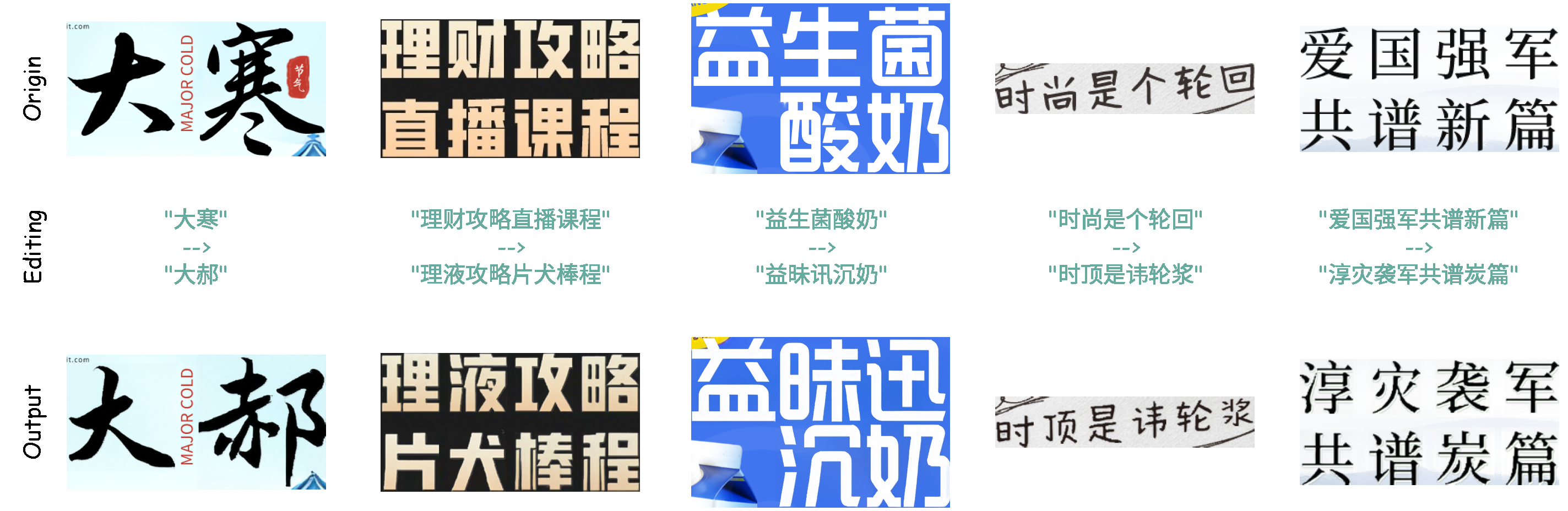}
\end{minipage}
\caption{More qualitative results for text editing in design images. The first row shows the original text region of the design images, the second row represents the edited text and the last row shows the edited outputs.}
\label{fig:comp-edit1}
\end{figure*}

\begin{figure*}[!t]
\begin{minipage}[!t]{0.8\linewidth}
\centering
\includegraphics[width=1\textwidth]{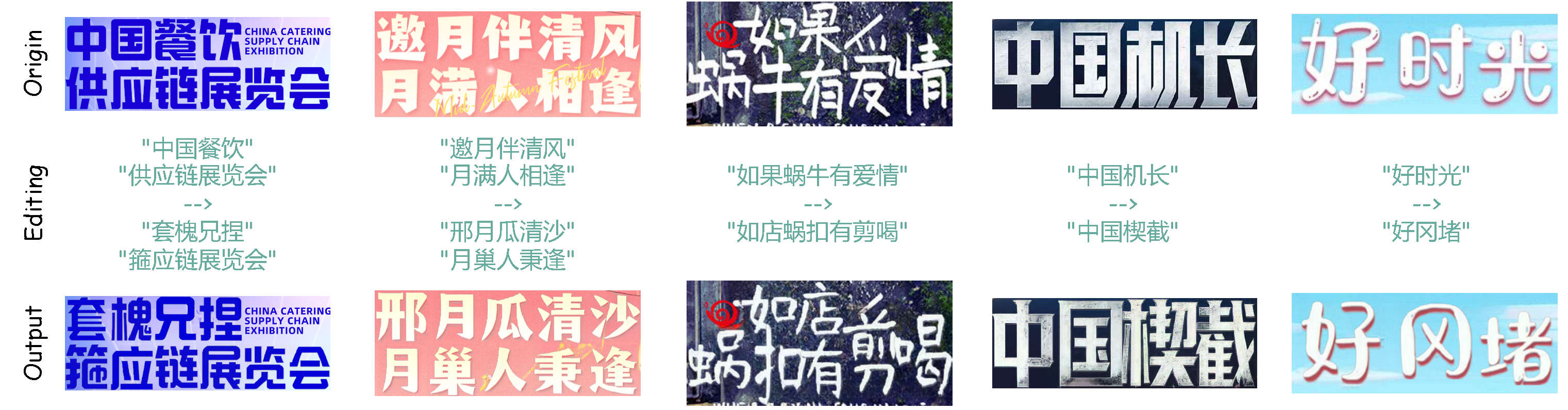}
\end{minipage}
\caption{More qualitative results for stylized text editing in design images. The first row shows the original text region of the design images, the second row represents the edited text and the last row shows the edited outputs.}
\label{fig:comp-edit2}
\end{figure*}

\begin{figure*}[!t]
\begin{minipage}[!t]{0.8\linewidth}
\centering
\includegraphics[width=1\textwidth]{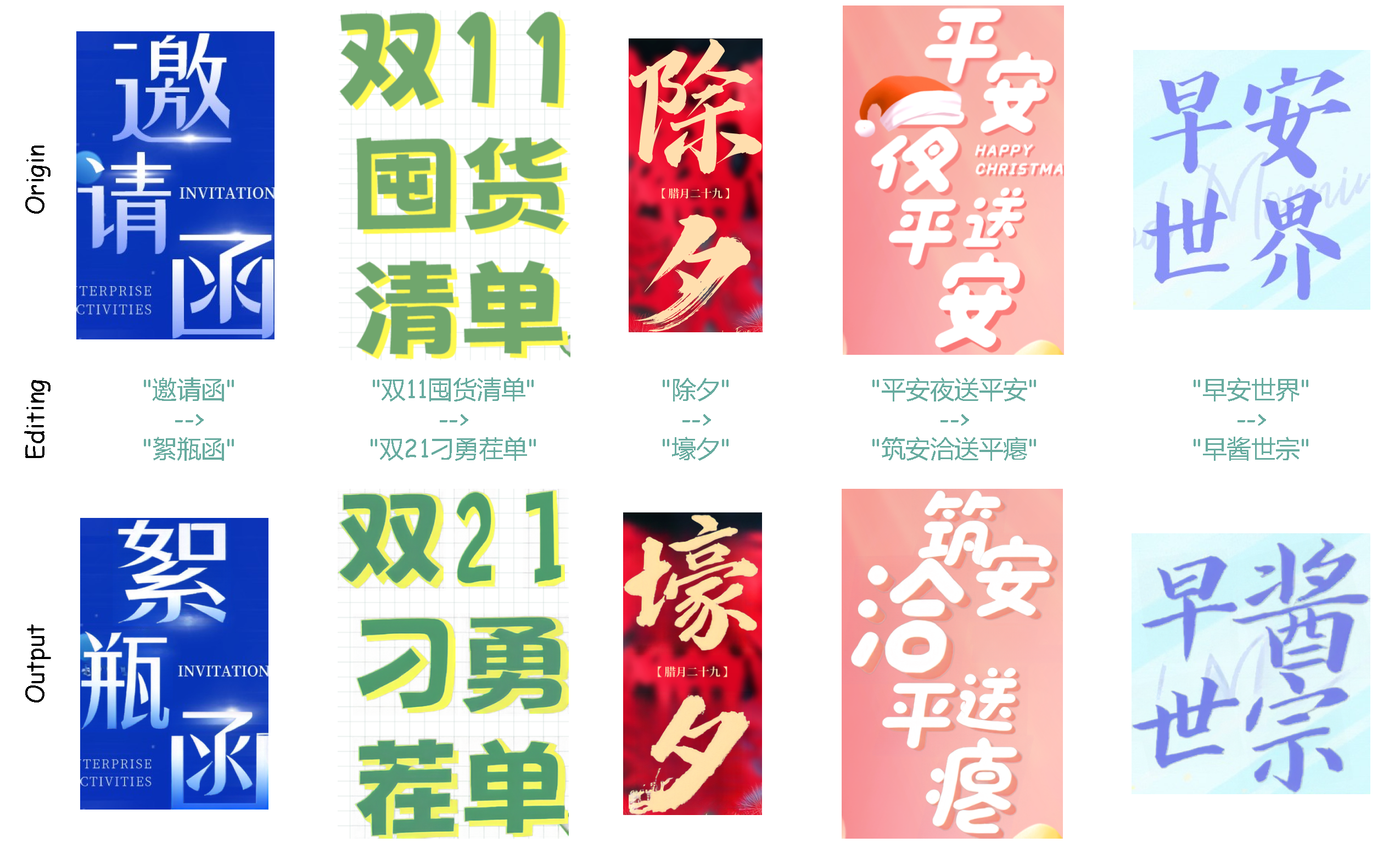}
\end{minipage}
\caption{More qualitative results for stylized text editing in design images. The first row shows the original text region of the design images, the second row represents the edited text and the last row shows the edited outputs.}
\label{fig:comp-edit3}
\end{figure*}

In this section, we present additional qualitative results of our method, covering two different tasks: stylized text editing in design images and full design image generation. As shown in~\cref{fig:comp-edit1},~\cref{fig:comp-edit2} and~\cref{fig:comp-edit3}, our approach enables arbitrary editing of stylized glyphs across a variety of scenarios such as posters and advertisements, while preserving the original background quality to the greatest extent and ensuring high fidelity and style consistency in the edited glyphs.

\begin{figure*}[!t]
\begin{minipage}[!t]{0.9\linewidth}
\centering
\includegraphics[width=1\textwidth]{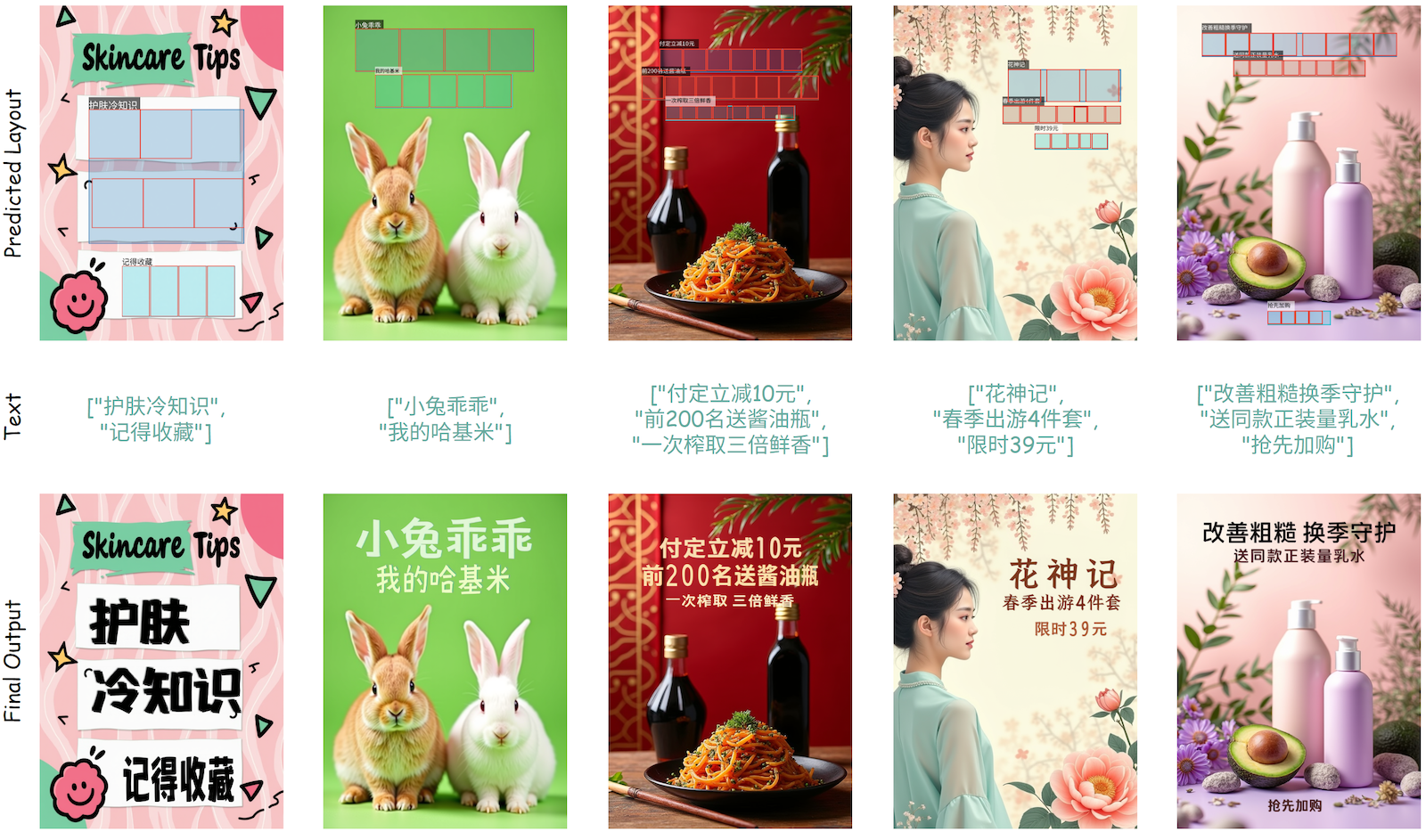}
\end{minipage}
\caption{More qualitative results for full design image generation. The first row shows the generated background and visualizes the predicted layout, the second row represents the text lines to render and the last row shows the final output.}
\label{fig:comp-system1}
\end{figure*}

\begin{figure*}[!t]
\begin{minipage}[!t]{0.9\linewidth}
\centering
\includegraphics[width=1\textwidth]{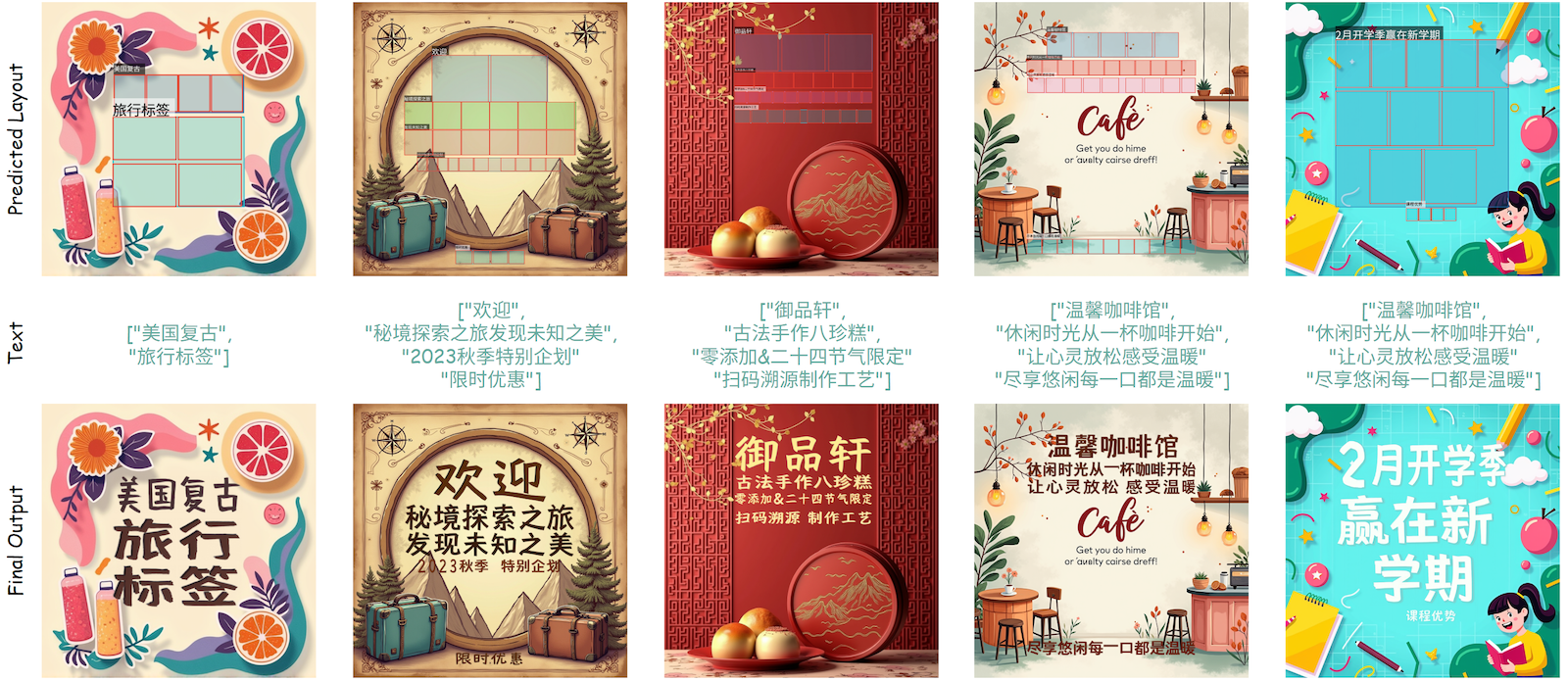}
\end{minipage}
\caption{More qualitative results for full design image generation. The first row shows the generated background and visualizes the predicted layout, the second row represents the text lines to render and the last row shows the final output.}
\label{fig:comp-system2}
\end{figure*}

\begin{figure*}[!t]
\begin{minipage}[!t]{0.9\linewidth}
\centering
\includegraphics[width=1\textwidth]{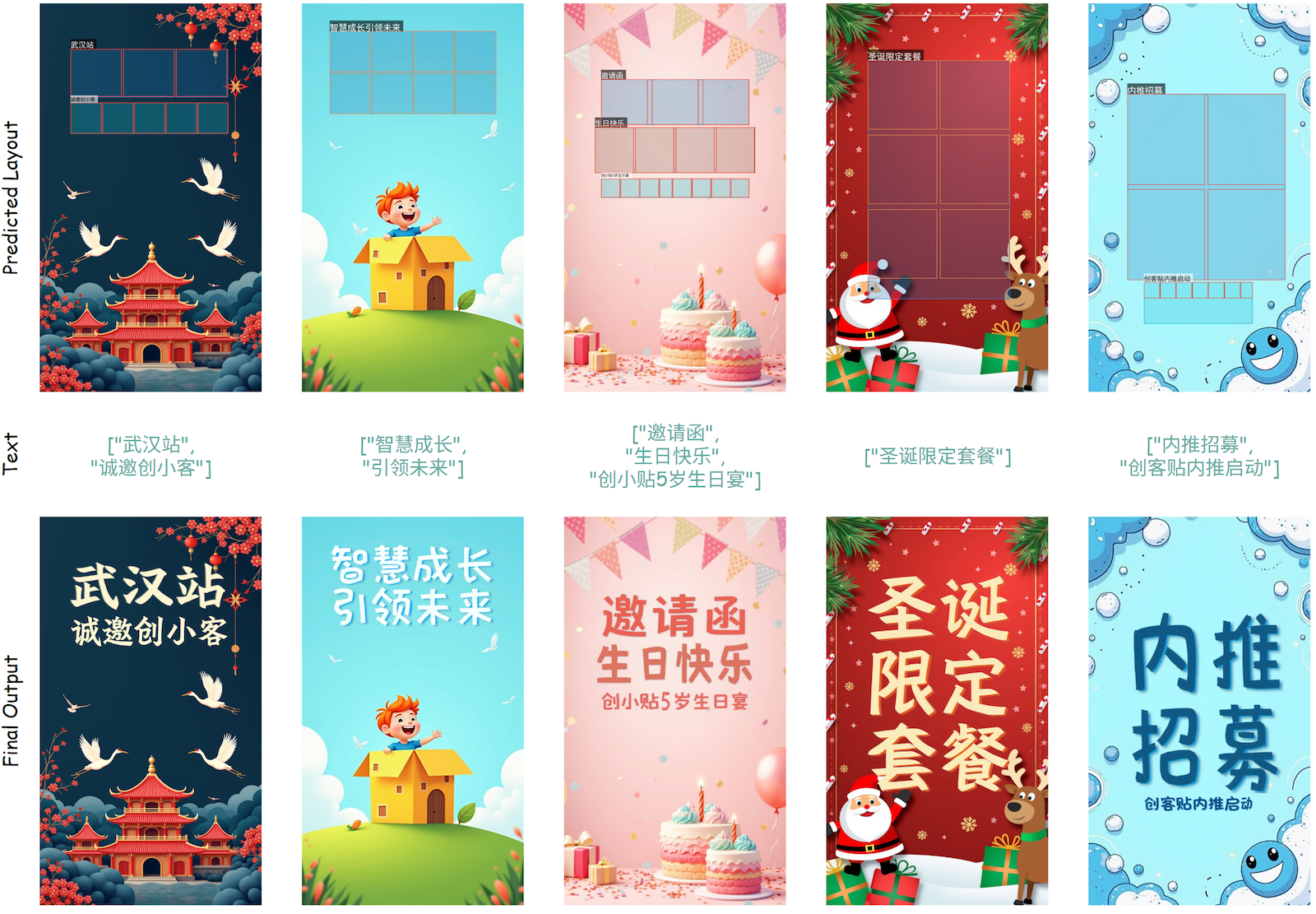}
\end{minipage}
\caption{More qualitative results for full design image generation. The first row shows the generated background and visualizes the predicted layout, the second row represents the text lines to render and the last row shows the final output.}
\label{fig:comp-system3}
\end{figure*}

As illustrated in~\cref{fig:comp-system1},~\cref{fig:comp-system2} and~\cref{fig:comp-system3}, our method is also capable of generating complete design images, including high-quality background synthesis, layout planning for foreground content, and accurate, style-consistent glyph rendering. Additionally, the prompts and rendered texts used for full design image generation in the main paper are listed in~\cref{tab:prompts} for reference.

\begin{figure*}[!t]
\begin{minipage}[!t]{0.8\linewidth}
\centering
\includegraphics[width=1\textwidth]{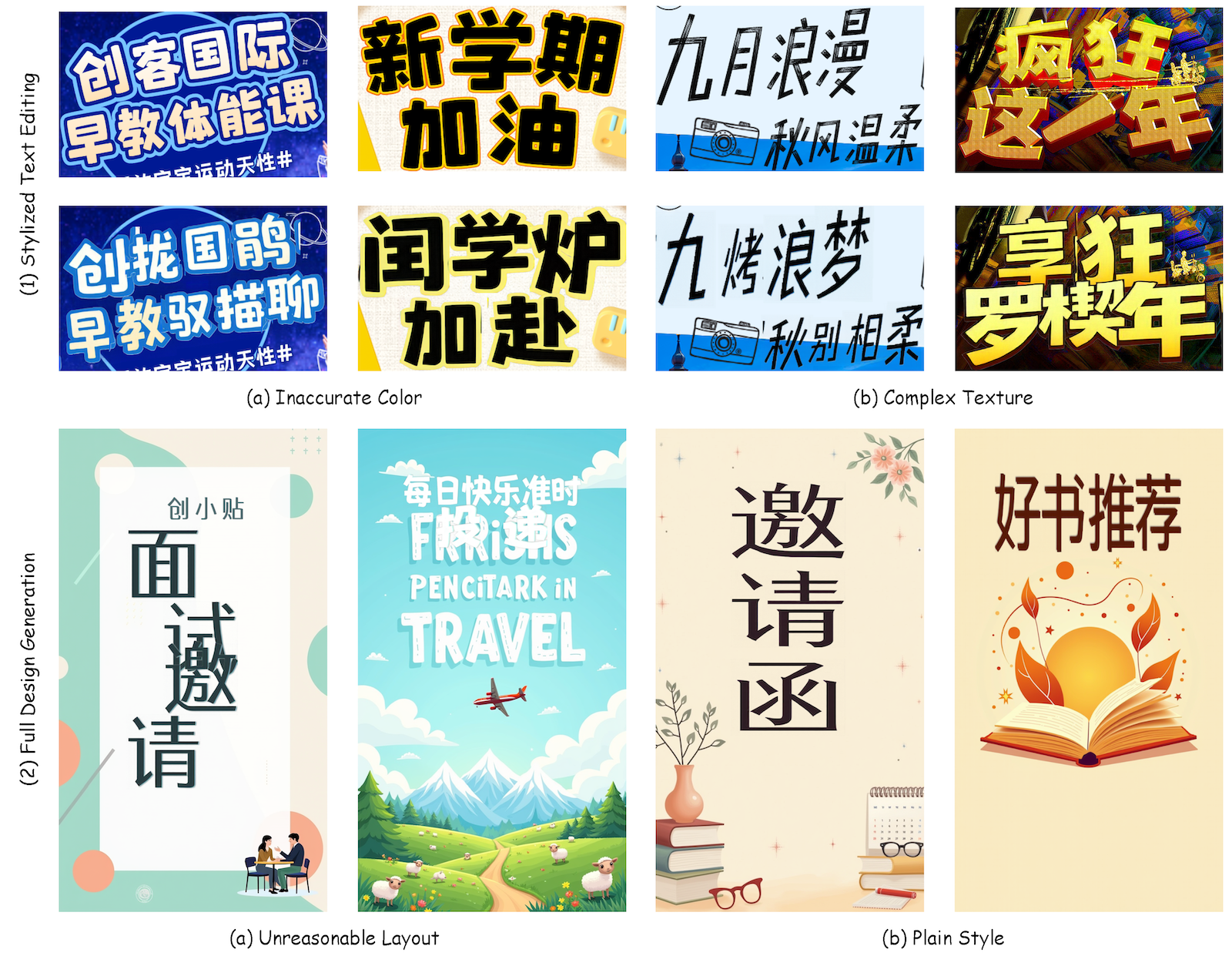}
\end{minipage}
\caption{Failure cases of \textbf{UTDesign}. We show failure cases for both stylized text editing (1) and full design generation (2).}
\label{fig:failure}
\end{figure*}

In addition, we provide some failure cases, as shown in~\cref{fig:failure}. For text editing, our method may occasionally fail to accurately transfer the target text's color (1a) and often struggles to handle overly complex or fine-grained font textures (1b). In the context of full design generation, the layout planner may sometimes produce suboptimal arrangements that lead to interference between foreground and background elements or between glyphs themselves (2a). In some cases, our glyph generation model may outputs overly plain styles, which can negatively affect the overall visual appeal (2b).

\section{More Details of the User Study}
\begin{figure*}[!t]
\begin{minipage}[!t]{0.8\linewidth}
\centering
\includegraphics[width=1\textwidth]{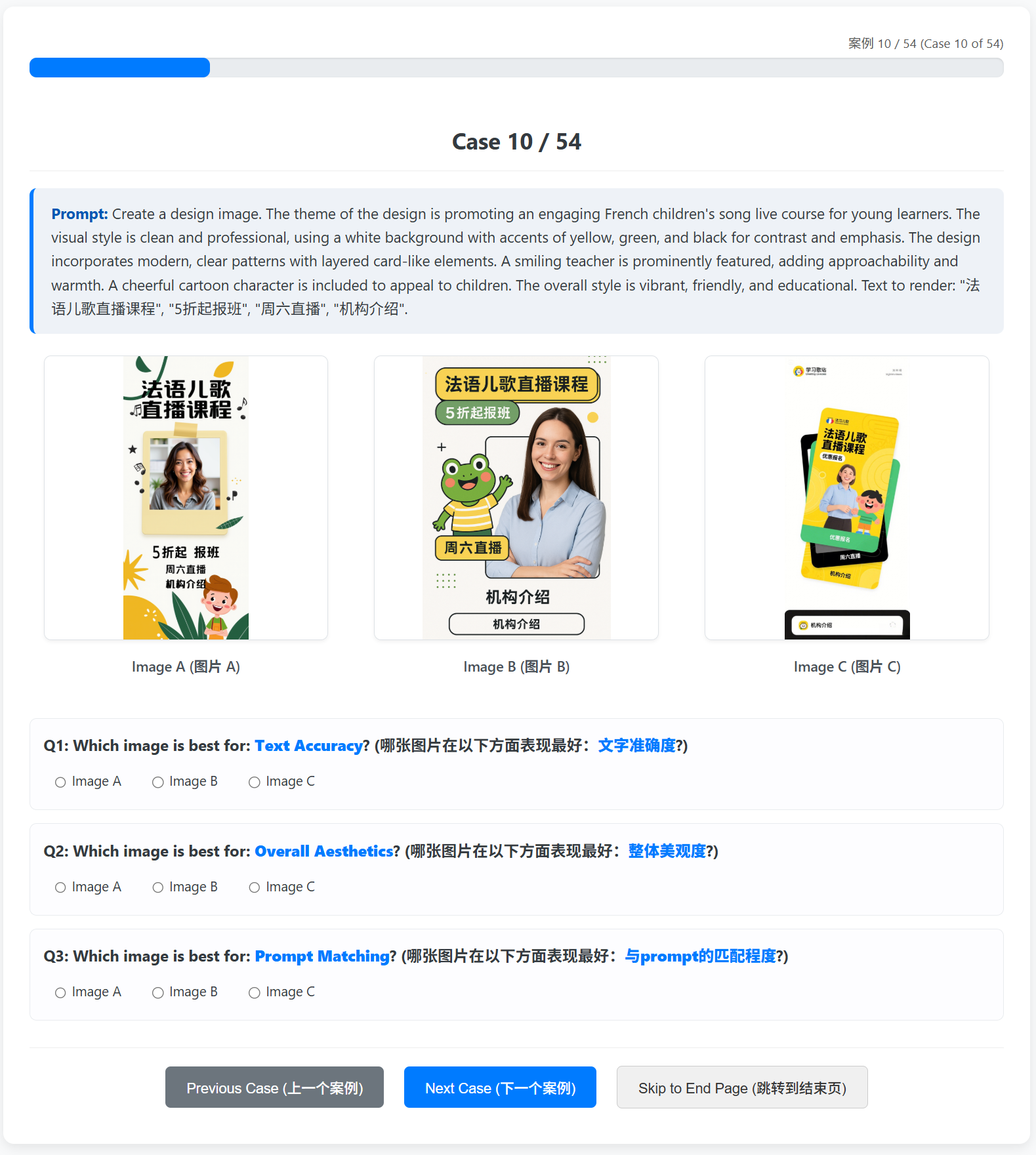}
\end{minipage}
\caption{Example of the interactive interface of our user study. The layout of each evaluation case is organized from top to bottom as follows: the prompt used in the current case, the image generation results from the three methods, and three questions used to assess different aspects of the outputs. In this case, image A is the output of our \textbf{UTDesign}.}
\label{fig:user page}
\end{figure*}

To compare the performance of our method with state-of-the-art proprietary approaches in design image generation, we first prepared 54 samples, including 43 selected from the \textbf{UTDesign-Bench-Gen} and 11 generated by a LLM using in-context learning (ICL). Each sample consists of a prompt and text rendered within a design. For each sample, we generated three images using different methods. These images were randomly shuffled and presented on an interactive webpage, where users were asked to select the best-performing one based on three evaluation criteria. An example of the interactive interface is shown in~\cref{fig:user page}. Our user study involved over 20 participants with diverse educational backgrounds and levels of expertise. We report the final results as the average win rate across the three evaluation aspects.

\section{{Ablation study on the base MLLM}}

\begin{table}[htbp]
\centering
\caption{Ablations on the base MLLM.}
\label{tab: abl-mllm}
\small
\resizebox{\linewidth}{!}{
\begin{tabular}{l|ccc|ccc}
\toprule
\multirow{2}{*}{\textbf{Settings}} & \multicolumn{3}{c|}{\textbf{Coarse Planning}} & \multicolumn{3}{c}{\textbf{Fine-grained Planning}} \\
\cmidrule(lr){2-4} \cmidrule(lr){5-7}
 & $\mathcal{R}_{iou}\uparrow$ & $-\mathcal{R}_{ol}\downarrow$ & FID$\downarrow$ & $\mathcal{R}_{iou}\uparrow$ & $-\mathcal{R}_{ol}\downarrow$ & $-\mathcal{R}_{bl}\downarrow$ \\
\midrule
Qwen2.5-VL-3B & 0.4396  & 0.0133  & 9.565  & 0.6501  & 0.0046  & 0.1895 \\
Qwen2.5-VL-7B & \textbf{0.5219} & 0.0071 & \textbf{8.146} & \textbf{0.6873} & \textbf{0.0024} & \textbf{0.0912} \\
Intern3-VL-8B & 0.2695  & 0.0071  & 13.65  &  0.6202 &  0.0065 &  0.2100 \\
\bottomrule
\end{tabular}}
\end{table}

Our editing system depends a lot on a good MLLM model. To validate the impact of base MLLM performance on the final text editing and layout planning results, we conduct an extensive ablation study across multiple MLLMs, comparing different base models and different model sizes. The results can be seen in~\cref{tab: abl-mllm}.

\begin{table*}[!t]
    \centering
    \caption{Prompts and text used for full design image generation in the paper.}
    \resizebox{0.95\linewidth}{!}{
    \begin{tabular}{|m{0.1\textwidth}|m{0.5\textwidth}|m{0.15\textwidth}|m{0.15\textwidth}|}
        \hline
        \small\textbf{Case} & \small\textbf{Prompt} & \small\textbf{Text to Render} & \small\textbf{Translated} \\
        \hline
        Row 1 
        & {\footnotesize The theme of the design is skincare tips, with a focus on providing easy-to-follow advice for better skin health. The design uses bright and vibrant colors like pink, green, and black, with playful and informal elements. The background features wavy pink patterns, and icons like arrows and a cartoon face add a dynamic and fun vibe. The layout is structured but lively with contrasting text boxes and symbols, giving it an engaging and approachable style.}
        & \begin{CJK}{UTF8}{gbsn}["护肤冷知识", "记得收藏"]\end{CJK}
        & ["Little-known Skincare Facts", "Remember to Save"] \\
        \hline
        Row 2
        & {\footnotesize The design theme centers on the sensation of summer heat, capturing the essence of warmth and relaxation. Visually, it features a light blue background suggesting a cool, airy atmosphere. Overlapping are hand-drawn white doodles and symbols, reminiscent of heat waves. A splash of bright yellow breaks the cool color palette, symbolizing the sun or intense heat, adding contrast and visual interest. The overall style is minimalistic and playful, capturing the whimsical and carefree spirit of summer.}
        & \begin{CJK}{UTF8}{gbsn}["夏天哪里都好", "就是太热热热", "热得只想躺"]\end{CJK}
        & ["Summer is great everywhere", "It's just hot, hot, hot", "So hot that I just want to lie down"] \\
        \hline
        Row 3
        & {\footnotesize The theme of the design is graduation celebration, conveying excitement and nostalgia for the milestone. The visual features a pastel gradient background transitioning from pink to green, with a clean, grid-like texture. It incorporates bold black elements alongside soft pink, green, and white accents, giving it a playful yet modern style. Cartoonish illustrations and minimalistic patterns, such as lines and borders, add a lively and youthful vibe to the overall design.}
        & \begin{CJK}{UTF8}{gbsn}["顶峰相见", "同学们", "热毕业啦", 不说再见"]\end{CJK}
        & ["Meet at the summit", "Classmates", "Graduated!", "Don't say goodbye"] \\
        \hline
        Row 4
        & {\footnotesize The design focuses on promoting travel during holidays with a cheerful and informative theme, highlighting cultural exploration. The visual features a pastel grid background resembling a notebook, accented with warm colors like beige and light blue. The central image showcases historical architecture surrounded by greenery, emphasizing cultural heritage. Decorative elements include a cartoon-style goddess, vehicle icons, and playful illustrations of landmarks. The overall style is light, organized, and inviting.}
        & \begin{CJK}{UTF8}{gbsn}["假期旅行攻略", "不踩雷的行程安排", "建议收藏"]\end{CJK}
        & ["Holiday Travel Guide", "Itinerary Without Pitfalls", "Suggested for Collection"] \\
        \hline
    \end{tabular}}
    \label{tab:prompts}
\end{table*}

\end{appendix}

%% file: main.bib
@String{Computer = "{IEEE} Computer" }

@String{Springer = "Springer-Verlag" }

@article{hu2024amo,
  title={AMO Sampler: Enhancing Text Rendering with Overshooting},
  author={Hu, Xixi and Xu, Keyang and Liu, Bo and Liu, Qiang and Fei, Hongliang},
  journal={arXiv preprint arXiv:2411.19415},
  year={2024}
}

@inproceedings{tuo2024anytext,
  title={AnyText: Multilingual Visual Text Generation and Editing},
  author={Tuo, Yuxiang and Xiang, Wangmeng and He, Jun-Yan and Geng, Yifeng and Xie, Xuansong},
  booktitle={ICLR},
  year={2024}
}

@article{tuo2024anytext2,
  title={AnyText2: Visual Text Generation and Editing With Customizable Attributes},
  author={Tuo, Yuxiang and Geng, Yifeng and Bo, Liefeng},
  journal={arXiv preprint arXiv:2411.15245},
  year={2024}
}

@inproceedings{zhang2024brush,
  title={Brush your text: Synthesize any scene text on images via diffusion model},
  author={Zhang, Lingjun and Chen, Xinyuan and Wang, Yaohui and Lu, Yue and Qiao, Yu},
  booktitle={Proceedings of the AAAI Conference on Artificial Intelligence},
  volume={38},
  number={7},
  pages={7215--7223},
  year={2024}
}

@article{wang2025designdiffusion,
  title={DesignDiffusion: High-Quality Text-to-Design Image Generation with Diffusion Models},
  author={Wang, Zhendong and Bao, Jianmin and Gu, Shuyang and Chen, Dong and Zhou, Wengang and Li, Houqiang},
  journal={arXiv preprint arXiv:2503.01645},
  year={2025}
}

@article{liu2024glyphv2,
  title={Glyph-byt5-v2: A strong aesthetic baseline for accurate multilingual visual text rendering},
  author={Liu, Zeyu and Liang, Weicong and Zhao, Yiming and Chen, Bohan and Liang, Lin and Wang, Lijuan and Li, Ji and Yuan, Yuhui},
  journal={arXiv preprint arXiv:2406.10208},
  year={2024}
}

@inproceedings{liu2024glyph,
  title={Glyph-byt5: A customized text encoder for accurate visual text rendering},
  author={Liu, Zeyu and Liang, Weicong and Liang, Zhanhao and Luo, Chong and Li, Ji and Huang, Gao and Yuan, Yuhui},
  booktitle={European Conference on Computer Vision},
  pages={361--377},
  year={2024},
  organization={Springer}
}

@article{yang2023glyphcontrol,
  title={Glyphcontrol: glyph conditional control for visual text generation},
  author={Yang, Yukang and Gui, Dongnan and Yuan, Yuhui and Liang, Weicong and Ding, Haisong and Hu, Han and Chen, Kai},
  journal={Advances in Neural Information Processing Systems},
  volume={36},
  pages={44050--44066},
  year={2023}
}

@inproceedings{ma2025glyphdraw2,
  title={Glyphdraw2: Automatic generation of complex glyph posters with diffusion models and large language models},
  author={Ma, Jian and Deng, Yonglin and Chen, Chen and Du, Nanyang and Lu, Haonan and Yang, Zhenyu},
  booktitle={Proceedings of the AAAI Conference on Artificial Intelligence},
  volume={39},
  number={6},
  pages={5955--5963},
  year={2025}
}

@article{zhao2025lex,
  title={LeX-Art: Rethinking Text Generation via Scalable High-Quality Data Synthesis},
  author={Zhao, Shitian and Wu, Qilong and Li, Xinyue and Zhang, Bo and Li, Ming and Qin, Qi and Liu, Dongyang and Zhang, Kaipeng and Li, Hongsheng and Qiao, Yu and others},
  journal={arXiv preprint arXiv:2503.21749},
  year={2025}
}

@article{chen2025posta,
  title={Posta: A go-to framework for customized artistic poster generation},
  author={Chen, Haoyu and Xu, Xiaojie and Li, Wenbo and Ren, Jingjing and Ye, Tian and Liu, Songhua and Chen, Ying-Cong and Zhu, Lei and Wang, Xinchao},
  journal={arXiv preprint arXiv:2503.14908},
  year={2025}
}

@article{gao2025postermaker,
  title={PosterMaker: Towards High-Quality Product Poster Generation with Accurate Text Rendering},
  author={Gao, Yifan and Lin, Zihang and Liu, Chuanbin and Zhou, Min and Ge, Tiezheng and Zheng, Bo and Xie, Hongtao},
  journal={arXiv preprint arXiv:2504.06632},
  year={2025}
}

@article{gong2025seedream,
  title={Seedream 2.0: A native chinese-english bilingual image generation foundation model},
  author={Gong, Lixue and Hou, Xiaoxia and Li, Fanshi and Li, Liang and Lian, Xiaochen and Liu, Fei and Liu, Liyang and Liu, Wei and Lu, Wei and Shi, Yichun and others},
  journal={arXiv preprint arXiv:2503.07703},
  year={2025}
}

@article{du2025textcrafter,
  title={TextCrafter: Accurately Rendering Multiple Texts in Complex Visual Scenes},
  author={Du, Nikai and Chen, Zhennan and Chen, Zhizhou and Gao, Shan and Chen, Xi and Jiang, Zhengkai and Yang, Jian and Tai, Ying},
  journal={arXiv preprint arXiv:2503.23461},
  year={2025}
}

@inproceedings{chen2024textdiffuser,
  title={Textdiffuser-2: Unleashing the power of language models for text rendering},
  author={Chen, Jingye and Huang, Yupan and Lv, Tengchao and Cui, Lei and Chen, Qifeng and Wei, Furu},
  booktitle={European Conference on Computer Vision},
  pages={386--402},
  year={2024},
  organization={Springer}
}

@inproceedings{zhao2024udifftext,
  title={Udifftext: A unified framework for high-quality text synthesis in arbitrary images via character-aware diffusion models},
  author={Zhao, Yiming and Lian, Zhouhui},
  booktitle={European Conference on Computer Vision},
  pages={217--233},
  year={2024},
  organization={Springer}
}

@inproceedings{zhang2023adding,
  title={Adding conditional control to text-to-image diffusion models},
  author={Zhang, Lvmin and Rao, Anyi and Agrawala, Maneesh},
  booktitle={Proceedings of the IEEE/CVF international conference on computer vision},
  pages={3836--3847},
  year={2023}
}

@inproceedings{esser2024scaling,
  title={Scaling rectified flow transformers for high-resolution image synthesis},
  author={Esser, Patrick and Kulal, Sumith and Blattmann, Andreas and Entezari, Rahim and M{\"u}ller, Jonas and Saini, Harry and Levi, Yam and Lorenz, Dominik and Sauer, Axel and Boesel, Frederic and others},
  booktitle={Forty-first international conference on machine learning},
  year={2024}
}

@article{jia2023cole,
  title={COLE: A Hierarchical Generation Framework for Multi-Layered and Editable Graphic Design},
  author={Jia, Peidong and Li, Chenxuan and Yuan, Yuhui and Liu, Zeyu and Shen, Yichao and Chen, Bohan and Chen, Xingru and Zheng, Yinglin and Chen, Dong and Li, Ji and others},
  journal={arXiv preprint arXiv:2311.16974},
  year={2023}
}

@inproceedings{inoue2024opencole,
  title={OpenCOLE: Towards Reproducible Automatic Graphic Design Generation},
  author={Inoue, Naoto and Masui, Kento and Shimoda, Wataru and Yamaguchi, Kota},
  booktitle={Proceedings of the IEEE/CVF Conference on Computer Vision and Pattern Recognition},
  pages={8131--8135},
  year={2024}
}

@article{pu2025art,
  title={Art: Anonymous region transformer for variable multi-layer transparent image generation},
  author={Pu, Yifan and Zhao, Yiming and Tang, Zhicong and Yin, Ruihong and Ye, Haoxing and Yuan, Yuhui and Chen, Dong and Bao, Jianmin and Zhang, Sirui and Wang, Yanbin and others},
  journal={arXiv preprint arXiv:2502.18364},
  year={2025}
}

@inproceedings{lin2023autoposter,
  title={Autoposter: A highly automatic and content-aware design system for advertising poster generation},
  author={Lin, Jinpeng and Zhou, Min and Ma, Ye and Gao, Yifan and Fei, Chenxi and Chen, Yangjian and Yu, Zhang and Ge, Tiezheng},
  booktitle={Proceedings of the 31st ACM International Conference on Multimedia},
  pages={1250--1260},
  year={2023}
}

@inproceedings{wang2022aesthetic,
  title={Aesthetic text logo synthesis via content-aware layout inferring},
  author={Wang, Yizhi and Pu, Guo and Luo, Wenhan and Wang, Yexin and Xiong, Pengfei and Kang, Hongwen and Lian, Zhouhui},
  booktitle={Proceedings of the IEEE/CVF Conference on Computer Vision and Pattern Recognition},
  pages={2436--2445},
  year={2022}
}

@article{patnaik2025aesthetiq,
  title={AesthetiQ: Enhancing Graphic Layout Design via Aesthetic-Aware Preference Alignment of Multi-modal Large Language Models},
  author={Patnaik, Sohan and Jain, Rishabh and Krishnamurthy, Balaji and Sarkar, Mausoom},
  journal={arXiv preprint arXiv:2503.00591},
  year={2025}
}

@article{he2024gldesigner,
  title={GLDesigner: Leveraging Multi-Modal LLMs as Designer for Enhanced Aesthetic Text Glyph Layouts},
  author={He, Junwen and Wang, Yifan and Wang, Lijun and Lu, Huchuan and He, Jun-Yan and Li, Chenyang and Chen, Hanyuan and Lan, Jin-Peng and Luo, Bin and Geng, Yifeng},
  journal={arXiv preprint arXiv:2411.11435},
  year={2024}
}

@article{li2024cgb,
  title={CGB-DM: Content and Graphic Balance Layout Generation with Transformer-based Diffusion Model},
  author={Li, Yu and Chen, Yifan and Liu, Gongye and Wu, Jie and Yang, Yujiu},
  journal={arXiv preprint arXiv:2407.15233},
  year={2024}
}

@inproceedings{tanglayoutnuwa,
  title={LayoutNUWA: Revealing the Hidden Layout Expertise of Large Language Models},
  author={Tang, Zecheng and Wu, Chenfei and Li, Juntao and Duan, Nan},
  booktitle={The Twelfth International Conference on Learning Representations}
}

@article{lin2023layoutprompter,
  title={Layoutprompter: awaken the design ability of large language models},
  author={Lin, Jiawei and Guo, Jiaqi and Sun, Shizhao and Yang, Zijiang and Lou, Jian-Guang and Zhang, Dongmei},
  journal={Advances in Neural Information Processing Systems},
  volume={36},
  pages={43852--43879},
  year={2023}
}

@inproceedings{hsu2023posterlayout,
  title={Posterlayout: A new benchmark and approach for content-aware visual-textual presentation layout},
  author={Hsu, Hsiao Yuan and He, Xiangteng and Peng, Yuxin and Kong, Hao and Zhang, Qing},
  booktitle={Proceedings of the IEEE/CVF Conference on Computer Vision and Pattern Recognition},
  pages={6018--6026},
  year={2023}
}

@article{seol2024posterllama,
  title={Posterllama: Bridging design ability of langauge model to contents-aware layout generation},
  author={Seol, Jaejung and Kim, Seojun and Yoo, Jaejun},
  journal={arXiv preprint arXiv:2404.00995},
  year={2024}
}

@article{yang2024posterllava,
  title={Posterllava: Constructing a unified multi-modal layout generator with llm},
  author={Yang, Tao and Luo, Yingmin and Qi, Zhongang and Wu, Yang and Shan, Ying and Chen, Chang Wen},
  journal={arXiv preprint arXiv:2406.02884},
  year={2024}
}

@inproceedings{li2023relation,
  title={Relation-aware diffusion model for controllable poster layout generation},
  author={Li, Fengheng and Liu, An and Feng, Wei and Zhu, Honghe and Li, Yaoyu and Zhang, Zheng and Lv, Jingjing and Zhu, Xin and Shen, Junjie and Lin, Zhangang and others},
  booktitle={Proceedings of the 32nd ACM International Conference on Information and Knowledge Management},
  pages={1249--1258},
  year={2023}
}

@inproceedings{chen2024pixart,
  title={Pixart-$\sigma$: Weak-to-strong training of diffusion transformer for 4k text-to-image generation},
  author={Chen, Junsong and Ge, Chongjian and Xie, Enze and Wu, Yue and Yao, Lewei and Ren, Xiaozhe and Wang, Zhongdao and Luo, Ping and Lu, Huchuan and Li, Zhenguo},
  booktitle={European Conference on Computer Vision},
  pages={74--91},
  year={2024},
  organization={Springer}
}

@article{team2024kolors,
  title={Kolors: Effective training of diffusion model for photorealistic text-to-image synthesis},
  author={Team, K},
  journal={arXiv preprint},
  year={2024}
}

@inproceedings{zheng2024cogview3,
  title={Cogview3: Finer and faster text-to-image generation via relay diffusion},
  author={Zheng, Wendi and Teng, Jiayan and Yang, Zhuoyi and Wang, Weihan and Chen, Jidong and Gu, Xiaotao and Dong, Yuxiao and Ding, Ming and Tang, Jie},
  booktitle={European Conference on Computer Vision},
  pages={1--22},
  year={2024},
  organization={Springer}
}

@article{liu2024playground,
  title={Playground v3: Improving text-to-image alignment with deep-fusion large language models},
  author={Liu, Bingchen and Akhgari, Ehsan and Visheratin, Alexander and Kamko, Aleks and Xu, Linmiao and Shrirao, Shivam and Lambert, Chase and Souza, Joao and Doshi, Suhail and Li, Daiqing},
  journal={arXiv preprint arXiv:2409.10695},
  year={2024}
}

@inproceedings{peebles2023scalable,
  title={Scalable diffusion models with transformers},
  author={Peebles, William and Xie, Saining},
  booktitle={Proceedings of the IEEE/CVF international conference on computer vision},
  pages={4195--4205},
  year={2023}
}

@article{liu2023visual,
  title={Visual instruction tuning},
  author={Liu, Haotian and Li, Chunyuan and Wu, Qingyang and Lee, Yong Jae},
  journal={Advances in neural information processing systems},
  volume={36},
  pages={34892--34916},
  year={2023}
}

@article{li2024llava,
  title={Llava-next-interleave: Tackling multi-image, video, and 3d in large multimodal models},
  author={Li, Feng and Zhang, Renrui and Zhang, Hao and Zhang, Yuanhan and Li, Bo and Li, Wei and Ma, Zejun and Li, Chunyuan},
  journal={arXiv preprint arXiv:2407.07895},
  year={2024}
}

@article{bai2023qwen,
  title={Qwen technical report},
  author={Bai, Jinze and Bai, Shuai and Chu, Yunfei and Cui, Zeyu and Dang, Kai and Deng, Xiaodong and Fan, Yang and Ge, Wenbin and Han, Yu and Huang, Fei and others},
  journal={arXiv preprint arXiv:2309.16609},
  year={2023}
}

@article{wang2024qwen2,
  title={Qwen2-vl: Enhancing vision-language model's perception of the world at any resolution},
  author={Wang, Peng and Bai, Shuai and Tan, Sinan and Wang, Shijie and Fan, Zhihao and Bai, Jinze and Chen, Keqin and Liu, Xuejing and Wang, Jialin and Ge, Wenbin and others},
  journal={arXiv preprint arXiv:2409.12191},
  year={2024}
}

@article{bai2025qwen2,
  title={Qwen2. 5-vl technical report},
  author={Bai, Shuai and Chen, Keqin and Liu, Xuejing and Wang, Jialin and Ge, Wenbin and Song, Sibo and Dang, Kai and Wang, Peng and Wang, Shijie and Tang, Jun and others},
  journal={arXiv preprint arXiv:2502.13923},
  year={2025}
}

@inproceedings{chen2024internvl,
  title={Internvl: Scaling up vision foundation models and aligning for generic visual-linguistic tasks},
  author={Chen, Zhe and Wu, Jiannan and Wang, Wenhai and Su, Weijie and Chen, Guo and Xing, Sen and Zhong, Muyan and Zhang, Qinglong and Zhu, Xizhou and Lu, Lewei and others},
  booktitle={Proceedings of the IEEE/CVF conference on computer vision and pattern recognition},
  pages={24185--24198},
  year={2024}
}

@article{chen2024expanding,
  title={Expanding performance boundaries of open-source multimodal models with model, data, and test-time scaling},
  author={Chen, Zhe and Wang, Weiyun and Cao, Yue and Liu, Yangzhou and Gao, Zhangwei and Cui, Erfei and Zhu, Jinguo and Ye, Shenglong and Tian, Hao and Liu, Zhaoyang and others},
  journal={arXiv preprint arXiv:2412.05271},
  year={2024}
}

@article{chen2023diffute,
  title={Diffute: Universal text editing diffusion model},
  author={Chen, Haoxing and Xu, Zhuoer and Gu, Zhangxuan and Li, Yaohui and Meng, Changhua and Zhu, Huijia and Wang, Weiqiang and others},
  journal={Advances in Neural Information Processing Systems},
  volume={36},
  pages={63062--63074},
  year={2023}
}

@inproceedings{radford2021learning,
  title={Learning transferable visual models from natural language supervision},
  author={Radford, Alec and Kim, Jong Wook and Hallacy, Chris and Ramesh, Aditya and Goh, Gabriel and Agarwal, Sandhini and Sastry, Girish and Askell, Amanda and Mishkin, Pamela and Clark, Jack and others},
  booktitle={International conference on machine learning},
  pages={8748--8763},
  year={2021},
  organization={PmLR}
}

@article{oquab2024dinov2,
  title={DINOv2: Learning Robust Visual Features without Supervision},
  author={Oquab, Maxime and Darcet, Timoth{\'e}e and Moutakanni, Th{\'e}o and Vo, Huy and Szafraniec, Marc and Khalidov, Vasil and Fernandez, Pierre and Haziza, Daniel and Massa, Francisco and El-Nouby, Alaaeldin and others},
  journal={Transactions on Machine Learning Research Journal},
  pages={1--31},
  year={2024}
}

@article{alayrac2022flamingo,
  title={Flamingo: a visual language model for few-shot learning},
  author={Alayrac, Jean-Baptiste and Donahue, Jeff and Luc, Pauline and Miech, Antoine and Barr, Iain and Hasson, Yana and Lenc, Karel and Mensch, Arthur and Millican, Katherine and Reynolds, Malcolm and others},
  journal={Advances in neural information processing systems},
  volume={35},
  pages={23716--23736},
  year={2022}
}

@article{kirstain2023pick,
  title={Pick-a-pic: An open dataset of user preferences for text-to-image generation},
  author={Kirstain, Yuval and Polyak, Adam and Singer, Uriel and Matiana, Shahbuland and Penna, Joe and Levy, Omer},
  journal={Advances in Neural Information Processing Systems},
  volume={36},
  pages={36652--36663},
  year={2023}
}

@inproceedings{wallace2024diffusion,
  title={Diffusion model alignment using direct preference optimization},
  author={Wallace, Bram and Dang, Meihua and Rafailov, Rafael and Zhou, Linqi and Lou, Aaron and Purushwalkam, Senthil and Ermon, Stefano and Xiong, Caiming and Joty, Shafiq and Naik, Nikhil},
  booktitle={Proceedings of the IEEE/CVF Conference on Computer Vision and Pattern Recognition},
  pages={8228--8238},
  year={2024}
}

@article{gao2025seedream3,
  title={Seedream 3.0 Technical Report},
  author={Gao, Yu and Gong, Lixue and Guo, Qiushan and Hou, Xiaoxia and Lai, Zhichao and Li, Fanshi and Li, Liang and Lian, Xiaochen and Liao, Chao and Liu, Liyang and others},
  journal={arXiv preprint arXiv:2504.11346},
  year={2025}
}

@article{zhang2024transparent,
  title={Transparent Image Layer Diffusion using Latent Transparency},
  author={Zhang, Lvmin and Agrawala, Maneesh},
  journal={ACM Transactions on Graphics (TOG)},
  volume={43},
  number={4},
  pages={1--15},
  year={2024},
  publisher={ACM New York, NY, USA}
}

@inproceedings{zhang2018unreasonable,
  title={The unreasonable effectiveness of deep features as a perceptual metric},
  author={Zhang, Richard and Isola, Phillip and Efros, Alexei A and Shechtman, Eli and Wang, Oliver},
  booktitle={Proceedings of the IEEE conference on computer vision and pattern recognition},
  pages={586--595},
  year={2018}
}

@article{heusel2017gans,
  title={Gans trained by a two time-scale update rule converge to a local nash equilibrium},
  author={Heusel, Martin and Ramsauer, Hubert and Unterthiner, Thomas and Nessler, Bernhard and Hochreiter, Sepp},
  journal={Advances in neural information processing systems},
  volume={30},
  year={2017}
}

@article{shao2024deepseekmath,
  title={Deepseekmath: Pushing the limits of mathematical reasoning in open language models},
  author={Shao, Zhihong and Wang, Peiyi and Zhu, Qihao and Xu, Runxin and Song, Junxiao and Bi, Xiao and Zhang, Haowei and Zhang, Mingchuan and Li, YK and Wu, Y and others},
  journal={arXiv preprint arXiv:2402.03300},
  year={2024}
}

@inproceedings{mishchenko2024prodigy,
  title={Prodigy: an expeditiously adaptive parameter-free learner},
  author={Mishchenko, Konstantin and Defazio, Aaron},
  booktitle={Proceedings of the 41st International Conference on Machine Learning},
  pages={35779--35804},
  year={2024}
}

@inproceedings{suvorov2022resolution,
  title={Resolution-robust large mask inpainting with fourier convolutions},
  author={Suvorov, Roman and Logacheva, Elizaveta and Mashikhin, Anton and Remizova, Anastasia and Ashukha, Arsenii and Silvestrov, Aleksei and Kong, Naejin and Goka, Harshith and Park, Kiwoong and Lempitsky, Victor},
  booktitle={Proceedings of the IEEE/CVF winter conference on applications of computer vision},
  pages={2149--2159},
  year={2022}
}

@inproceedings{kirillov2023segment,
  title={Segment anything},
  author={Kirillov, Alexander and Mintun, Eric and Ravi, Nikhila and Mao, Hanzi and Rolland, Chloe and Gustafson, Laura and Xiao, Tete and Whitehead, Spencer and Berg, Alexander C and Lo, Wan-Yen and others},
  booktitle={Proceedings of the IEEE/CVF international conference on computer vision},
  pages={4015--4026},
  year={2023}
}

@article{khanam2024yolov11,
  title={Yolov11: An overview of the key architectural enhancements},
  author={Khanam, Rahima and Hussain, Muhammad},
  journal={arXiv preprint arXiv:2410.17725},
  year={2024}
}
